\documentclass[lettersize,journal]{IEEEtran}

\usepackage[table]{xcolor}   % 支持 \rowcolor, \cellcolor
\usepackage{tabularx}        % 支持自动调整列宽的 tabularx 表格环境
\usepackage{booktabs}        % 用于 \toprule, \midrule 等
\usepackage{multirow}        % 多行单元格
\usepackage{amsmath}         % 数学符号和公式
\usepackage{makecell}

\usepackage{paralist}
\usepackage{amsfonts}
\usepackage{algorithmic}
\usepackage{algorithm}
\usepackage{array}
\usepackage{textcomp}
\usepackage{stfloats}
\usepackage{url}
\usepackage{verbatim}
\usepackage{graphicx}
\usepackage{cite}
\usepackage{hyperref}
\usepackage[hyphenbreaks]{breakurl}
\usepackage{siunitx} % for better number alignment
\ifCLASSOPTIONcompsoc
  \usepackage{subfig}
\else
  \usepackage[caption=false, font=footnotesize]{subfig}
\fi

\def \NOTE [#1]{\textcolor{blue}{(\textit{#1})}}
\usepackage{xspace}
\usepackage{color}
\usepackage{multirow}
\usepackage{ifthen}

\long\def\ignorethis#1{}

\definecolor{gray}{rgb}{0.35,0.35,0.35}
\definecolor{MyBlue}{rgb}{0,0.2,0.8}
\definecolor{MyRed}{rgb}{0.8,0.2,0}
\definecolor{MyGreen}{rgb}{0.0,0.5,0.1}
\definecolor{MyGray}{rgb}{0.4,0.4,0.4}

\newlength\paramargin
\newlength\figmargin
\newlength\subfigmargin
\newlength\secmargin
\newlength\subsecmargin
\newlength\tabmargin
\newlength\eqmargin

\usepackage{array}
\newcolumntype{L}[1]{>{\raggedright\let\newline\\\arraybackslash\hspace{0pt}}m{#1}}
\newcolumntype{C}[1]{>{\centering\let\newline\\\arraybackslash\hspace{0pt}}m{#1}}
\newcolumntype{R}[1]{>{\raggedleft\let\newline\\\arraybackslash\hspace{0pt}}m{#1}}

\def\ie{i.e.,~}
\def\eg{e.g.,~}

\newcommand{\secref}[1]{Section~\ref{sec:#1}}

\newcommand{\figref}[1]{Fig.~\ref{fig:#1}}
\newcommand{\tabref}[1]{Table~\ref{tab:#1}}

\newcommand{\Paragraph}[1]{\noindent\textbf{#1}}

\usepackage{pifont}
\usepackage{bbding}
\usepackage{float}
\usepackage{cite}
\definecolor{mycolor_blue}{RGB}{231,239,250}
\definecolor{mycolor_green}{RGB}{230,247,224}
\definecolor{mycolor_gray}{RGB}{236,236,236}
\definecolor{pearDark!20}{RGB}{212,230,241}

\newcolumntype{P}[1]{>{\raggedright\arraybackslash}p{#1\linewidth}} % 自动适应文本宽度
\begin{document}

\title{UniMIC: Token-Based Multimodal Interactive Coding for Human–AI Collaboration}
%
%
% author names and IEEE memberships
% note positions of commas and nonbreaking spaces ( ~ ) LaTeX will not break
% a structure at a ~ so this keeps an author's name from being broken across
% two lines.
% use \thanks{} to gain access to the first footnote area
% \def\etal{\textit{et~al}.\xspace}
% \def\ie{\textit{i.e.},\xspace}
\author{Qi~Mao, Tinghan~Yang, Jiahao~Li, Bin~Li, Libiao~Jin, Yan~Lu%
\thanks{Qi Mao, Tinghan Yang, and Libiao Jin are with the State Key Laboratory of Media Convergence and Communication, Communication University of China, Beijing 100024, China (e-mail: qimao@cuc.edu.cn; yangtinghan@cuc.edu.cn; libiao@cuc.edu.cn).}%
\thanks{Jiahao Li, Bin Li, and Yan Lu are with Microsoft Research Asia, Beijing 10080, China (e-mail: li.jiahao@microsoft.com; libin@microsoft.com; yanlu@microsoft.com).}
\thanks{This work was done when Qi Mao was a visiting scholar at Microsoft Research Asia.}
}

% The paper headers
\markboth{Under Review}%
{Qi Mao\MakeLowercase{\textit{et al.}}: A Sample Article Using IEEEtran.cls for IEEE Journals}

%\IEEEpubid{0000--0000/00\$00.00~\copyright~2021 IEEE}

% make the title area
\maketitle

% in the abstract or keywords.
\IEEEpeerreviewmaketitle
\begin{abstract}
The rapid progress of Large Multimodal Models (LMMs) and cloud-based AI agents is transforming human–AI collaboration into bidirectional, multimodal interaction. However, existing codecs remain optimized for unimodal, one-way communication, resulting in repeated degradation under conventional compress–transmit–reconstruct pipelines.
To address this limitation, we propose UniMIC, a \underline{Uni}fied token-based \underline{M}ultimodal \underline{I}nteractive \underline{C}oding framework that bridges edge devices and cloud AI agents. 
Instead of transmitting raw pixels or plain text, UniMIC employs compact tokenized representations as the communication medium, enabling efficient low-bitrate transmission while maintaining compatibility with LMMs. To further enhance compression, lightweight Transformer-based entropy models with scenario-specific designs—generic, masked, and text-conditioned—effectively minimize inter-token redundancy.
Extensive experiments on text-to-image generation, text-guided inpainting, outpainting, and visual question answering show that UniMIC achieves substantial bitrate savings and remains robust even at ultra-low bitrates ($<0.05$ bpp), without compromising downstream task performance. These results establish UniMIC as a practical and forward-looking paradigm for next-generation multimodal interactive communication.
% \iffalse
% \fi
\end{abstract}

\section{Introduction}
\IEEEPARstart{R}{}ecent advancements in artificial intelligence (AI), particularly Large Multimodal Models (LMMs) and autonomous AI agents, are fundamentally reshaping the paradigm of human--AI collaboration.  
As AI systems---typically deployed in the cloud---evolve from passive analytical tools to interactive collaborators (\eg generative design assistants that co-create content with users or diagnostic agents that conduct multimodal consultations), the interaction pattern between humans and machines is shifting from \textbf{one-way send--receive communication to iterative, bidirectional, and multimodal dialogue.} 
In such collaborative settings, \emph{edge devices} (\ie user terminals) transmit text instructions and upload images for analysis, while \emph{cloud-based AI agents} respond by generating new content (\eg text or images) or performing multimodal reasoning according to the given instructions.  
This process inherently involves both humans and AI as senders and receivers, with repeated multimodal exchanges over multiple rounds, thereby demanding a new communication paradigm for human--AI collaboration.  

\begin{figure}[!t]
    \centering \includegraphics[width=1.0\linewidth]{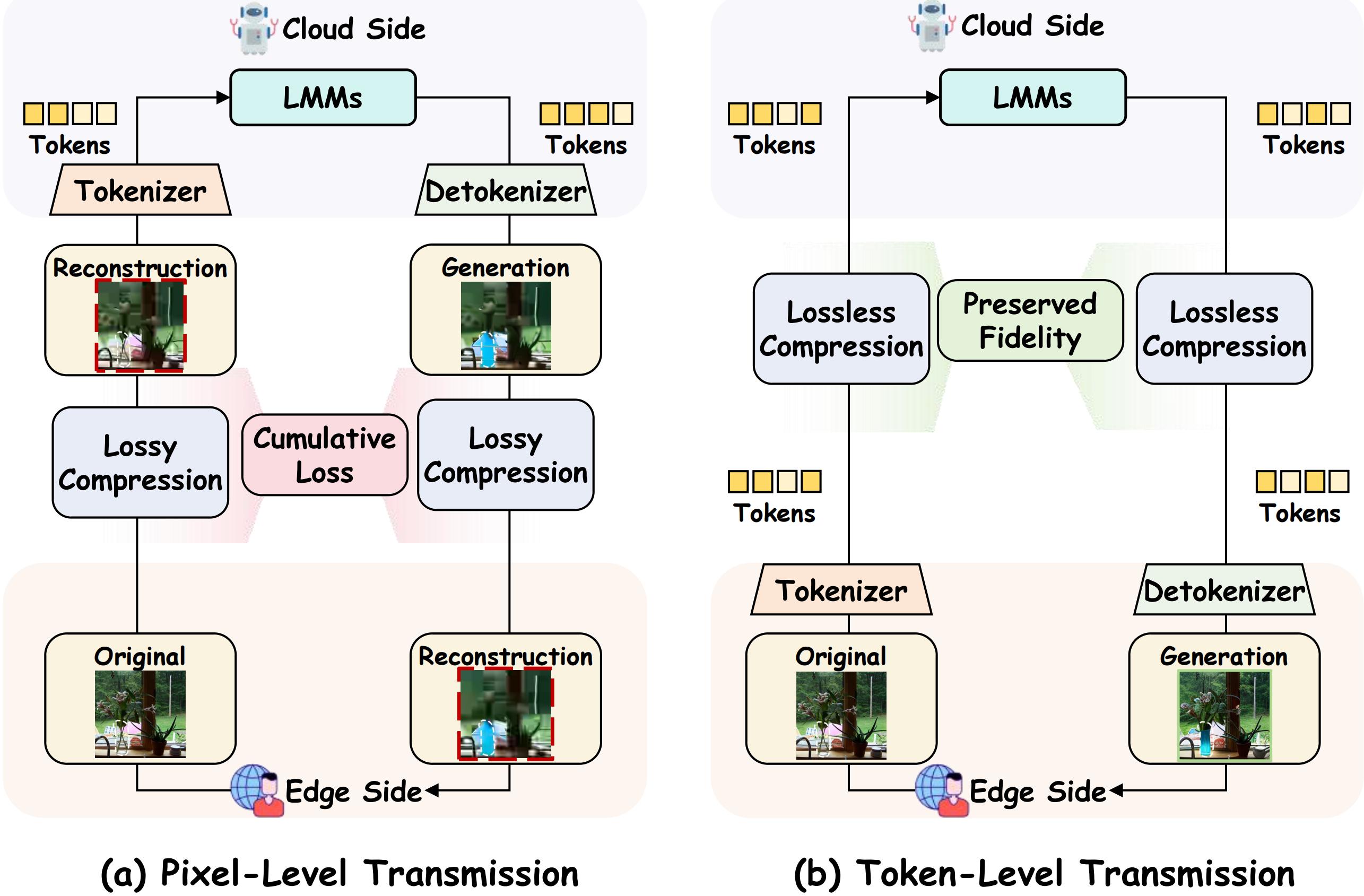}    \caption{\textbf{Comparison of pixel- and token-level transmission in human--AI interaction.} 
Illustrated with the inpainting task (text prompts omitted for simplicity). 
(a) Pixel-based pipelines accumulate distortion due to repeated image compression--decompression. 
(b) Token-based pipelines exchange losslessly compressed tokens, preserving fidelity even at ultra-low bitrates.}
\label{fig:teaser1}
    \vspace{-6 mm}
\end{figure}

However, existing compression paradigms are not well-suited to this interactive setting.  
Most traditional frameworks are designed either for human perception (\eg minimizing visual distortion)~\cite{pennebaker1992jpeg,rabbani2002jpeg2000,wiegand2003overview,gao2014overview,sullivan2012overview} or for machine understanding~\cite{duan2015overview,duan2018compact} in limited recognition tasks, but they typically treat humans and machines as isolated endpoints.  
Recent works~\cite{duan2020video,wang2021towards,hu2020towards,choi2022scalable,yan2021sssic,tu2021semantic,liu2021semantics} have begun to consider both humans and machines as receivers, yet these designs still assume \emph{unidirectional, unimodal} transmission and fail to support iterative multimodal interaction.  
As a result, conventional pipelines often compress raw pixels before sending them to cloud LLMs for processing, and then re-compress the generated results before delivering them back to users, causing repeated degradation and latency (see \figref{teaser1}(a)).

\begin{figure*}[!t]
    \centering
 \includegraphics[width=0.98\linewidth]{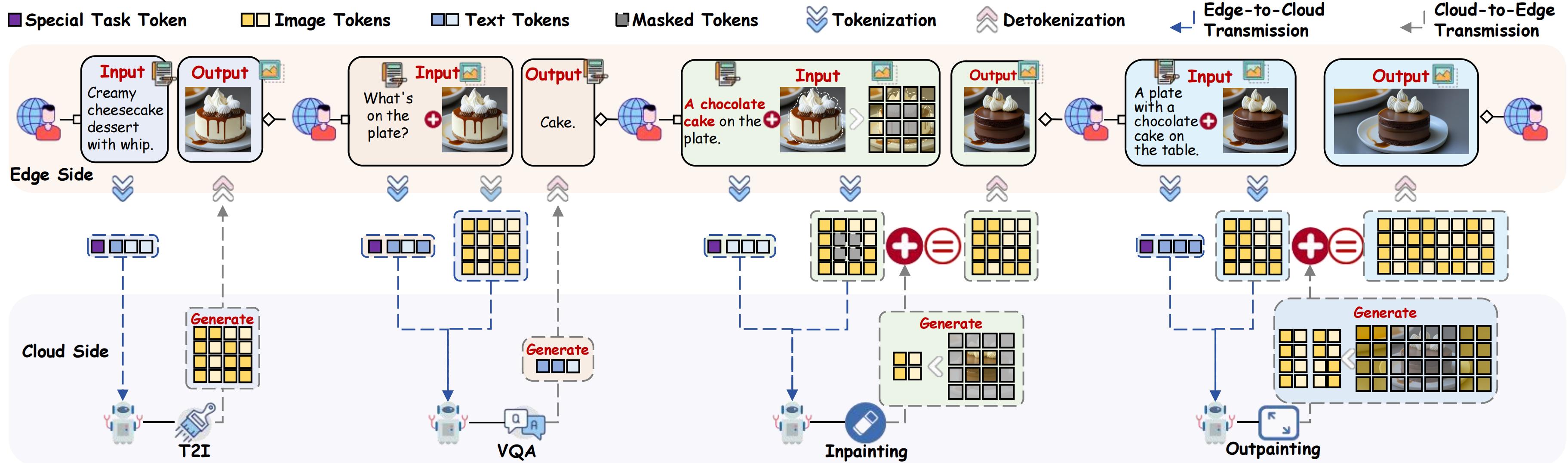}
 \vspace{-1 mm}
\caption{\textbf{Application scenarios of the proposed UniMIC framework.} 
All tasks share the same token-based transmission pipeline but transmit different token subsets, enabling efficient and flexible multimodal communication.
}
    \label{fig:applications}
    \vspace{-6 mm}
\end{figure*}

To address these limitations, we propose \textbf{\emph{UniMIC}}, a unified \emph{token-based multimodal interactive coding framework} tailored for human--AI collaboration.  
Different from existing codecs that only optimize rate--distortion in pixels or features, UniMIC establishes an \textbf{AI-native communication protocol} where \emph{tokens serve as the native medium of exchange}. 
This protocol enables edge devices and cloud agents to exchange only the task-relevant token subsets.
For instance, as illustrated in \figref{applications}, in inpainting, the edge transmits unmasked tokens and editing instructions, while the cloud returns only the generated tokens for masked regions; in outpainting, the edge uploads full tokens and the cloud transmits only extrapolated tokens back.
Such task-adaptive and asymmetric transmission is fundamentally different from traditional compress--transmit--reconstruct loops. 
UniMIC incurs only a one-time tokenization loss, while all subsequent exchanges are lossless, thereby avoiding cumulative degradation and preserving semantic fidelity even at ultra-low bitrates, as shown in \figref{teaser1}(b).

Building on this protocol, we further develop lightweight Transformer-based entropy models specialized for different scenarios, including autoregressive, masked-token, and text-conditional variants.  
These models reduce token redundancy and align probability estimation with generative statistics, ensuring efficiency across diverse multimodal tasks.  
Consequently, UniMIC supports a broad range of downstream applications, including text-to-image (T2I) generation, text-guided inpainting, outpainting, and visual question answering (VQA), with consistent gains in both compression efficiency and task performance.

The main contributions of this work are summarized as follows:
\begin{itemize}
    \item We establish \textbf{\emph{UniMIC}}, a unified token-based multimodal interactive coding framework that treats tokens as the native medium of exchange between edge devices and cloud AI agents. This paradigm enables bidirectional, task-adaptive communication while avoiding cumulative degradation inherent in pixel-based codecs.
    \item We develop a family of lightweight Transformer-based entropy models, including autoregressive, masked-token, and text-conditional variants, together with scenario-specific training strategies to minimize redundancy and adapt to diverse interaction settings.
    \item We extensively validate UniMIC on T2I generation, text-guided inpainting, outpainting, and VQA, showing consistent bitrate savings and improved task performance over existing pixel-based baselines at ultra-low bitrates ($<0.05$ bpp).
\end{itemize}

\section{Related work}
\label{sec:related work}

\subsection{Image/Video Compression} Classical image and video compression has evolved over several decades, resulting in standardized formats such as JPEG~\cite{pennebaker1992jpeg}, JPEG2000~\cite{rabbani2002jpeg2000}, MPEG-4, AVC/H.264~\cite{wiegand2003overview}, AVS~\cite{gao2014overview}, and HEVC~\cite{sullivan2012overview}. 
The latest standard, VVC~\cite{bross2021overview}, further improves rate--distortion efficiency through more flexible coding tools. 
However, classic codecs are built upon a modular pipeline in which each stage---including prediction, transform, quantization, and entropy coding---relies on handcrafted design choices. These modules are typically optimized in isolation, leading to locally optimal solutions that cannot fully exploit cross-module dependencies.

The emergence of deep learning has enabled a new paradigm in which the entire compression pipeline can be learned end-to-end. Deep neural network (DNN)-based codecs~\cite{theis2017lossy,cheng2020learned,kim2022joint} replace handcrafted transforms and entropy models with trainable components, jointly optimized for global rate--distortion objectives. 
This approach offers better adaptability to image content and supports flexible bitrate control~\cite{gao2022flexible}. However, most DNN-based codecs remain primarily distortion-oriented and may struggle to preserve perceptual quality at extremely low bitrates.

To address this, deep generative models have been integrated into learned compression frameworks. Generative adversarial networks (GANs)~\cite{goodfellow2014generative} excel at synthesizing high-resolution details from compact latent codes~\cite{agustsson2019generative,chang2019layered,chang2022conceptual}, while diffusion-based models further improve perceptual fidelity in ultra-low bitrate regimes~\cite{rippel2017real,lee2020training,iwai2021fidelity}. Nevertheless, both DNN- and generative-based codecs largely follow a compress-then-analyze (CTA) paradigm~\cite{redondi2013compress}, requiring full image reconstruction before any downstream visual analysis. 
This design introduces unnecessary decoding overhead and is suboptimal for machine-oriented applications. In contrast, our work departs from the CTA paradigm by leveraging token-based transmission to jointly support efficient machine analysis and high-quality human perception.

\subsection{Human–Machine Collaborative Compression}
As outlined in the previous subsection, conventional machine-oriented codecs often follow a CTA paradigm, which requires full image decoding before downstream analysis and is inefficient for many AI-driven applications.  
An alternative, the \emph{analyze-then-compress} (ATC) paradigm~\cite{redondi2013compress}, reduces transmission costs in machine vision tasks by transmitting task-specific features~\cite{duan2015overview}, but this approach severely limits reconstruction quality, making results unsuitable for human inspection.

To address this limitation, the \emph{Human–Machine Collaborative Compression} (HMCC) paradigm~\cite{duan2020video} jointly optimizes for both human and machine receivers. 
Existing works adopt scalable coding architectures that decompose content into semantic and perceptual layers. Examples include face-recognition-based residual enhancement~\cite{wang2021towards}, unified decoders for gradual reconstruction~\cite{yang2021towards}, task-specific latent partitions~\cite{choi2022scalable}, and semantics-driven hierarchical schemes for progressive decoding~\cite{yan2021sssic,tu2021semantic,mao2023scalable}.
% ~\cite{yan2021sssic,tu2021semantic,liu2021semantics,mao2023scalable}.

Despite these advances, two major challenges remain:
(1) current HMCC frameworks assume fixed sender–receiver roles, lacking flexibility for interactive bidirectional communication, in which both human and machine serve as sender and receiver;
(2) they largely focus on visual-only data, offering limited support for multimodal tasks that involve cross-modal reasoning or generation.
Our work addresses these gaps by introducing a discrete token-based coding framework that unifies multimodal information. 
This design supports flexible role reversals and enables efficient, high-quality reconstructions for both human and machine receivers in diverse human–AI collaborative tasks.

\begin{figure*}[!t]
    \centering
    % \vspace{-3mm}
    \includegraphics[width=1.0\linewidth]{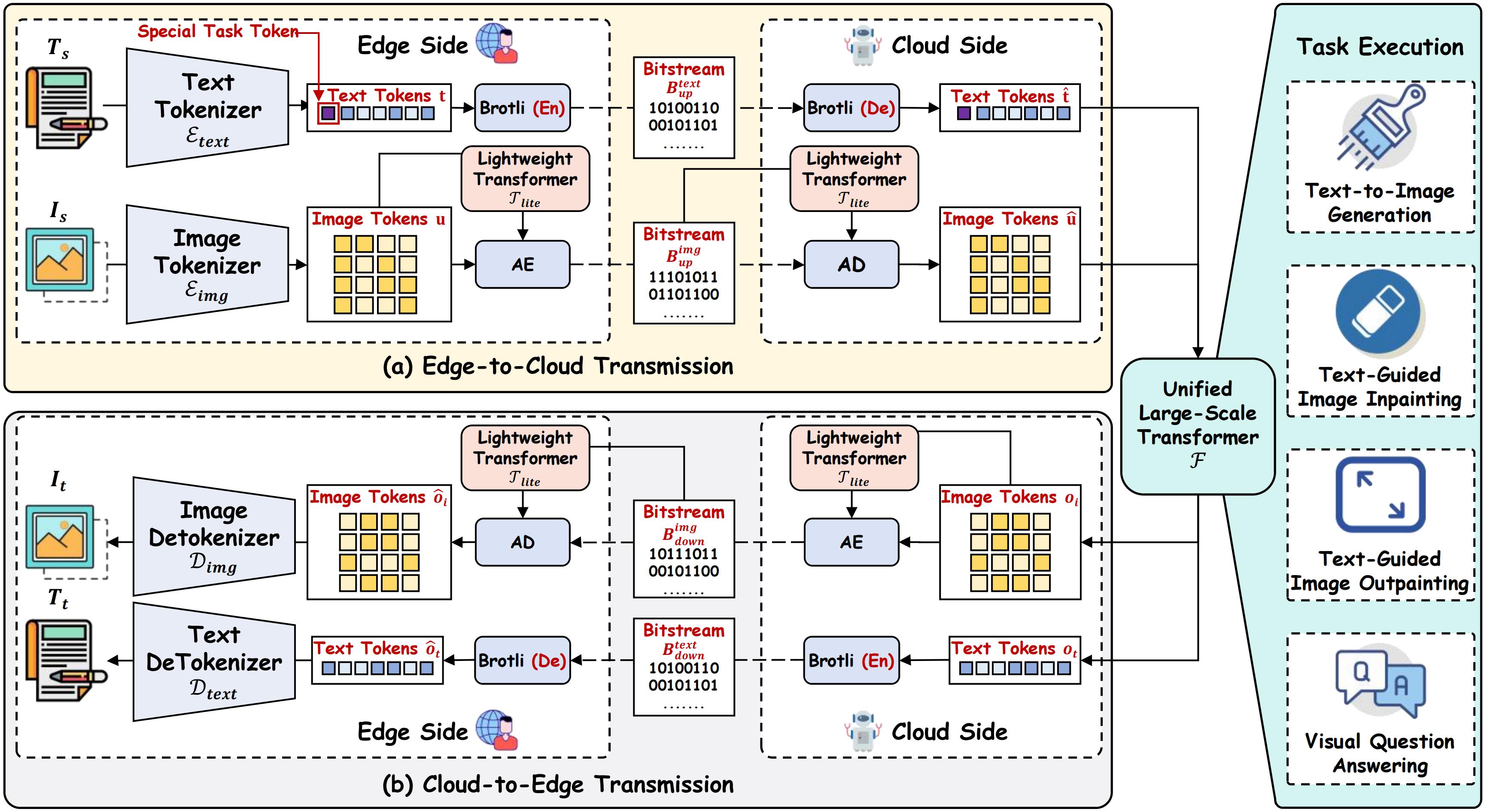}
\caption{\textbf{Overall architecture of the proposed UniMIC framework.} 
(a) In edge-to-cloud transmission, multimodal inputs are tokenized and entropy-coded before being sent to the cloud. 
(b) In cloud-to-edge transmission, the Unified Transformer processes the tokens, and the generated outputs are entropy-coded and reconstructed at the user side.
} 
    \vspace{- 5mm}
    \label{fig:framework}
\end{figure*}

\subsection{Large Multimodal Models}
LMMs have recently demonstrated unprecedented capabilities in both understanding and generation, enabled by unified representation learning across visual, textual, and auditory modalities~\cite{chowdhery2023palm,brown2020language,touvron2023llama}.  
By mapping heterogeneous inputs into a shared embedding space, they can seamlessly perform diverse cross-modal tasks—such as image captioning and visual question answering~\cite{liu2023llava,dai2023instructblip,wu2024next}—without relying on separate task-specific models.

Building on this foundation, recent works such as Chameleon~\cite{team2024chameleon}, Transfusion~\cite{zhou2024transfusion}, Show-o~\cite{zhang2024show} and UniTok~\cite{unitok} employ discrete tokenization to represent multimodal data streams within a unified vocabulary, enabling simultaneous comprehension and generation in Transformer-based architectures.  
This establishes discrete tokens as a fundamental unit for scalable multimodal reasoning and generation.

Motivated by these advances, we introduce the first multimodal \emph{interactive} coding framework that operates directly on LMM tokens.  
By compressing discrete token sequences—rather than raw pixels or text—we achieve low-bitrate transmission while fully preserving the semantic fidelity and generative capabilities of large cloud-based models, thus enabling efficient, high-quality human–AI collaboration.

\section{UniMIC: Token-Based Multimodal Interactive Coding}
\label{sec:proposed-method}

We present \textbf{UniMIC}, an efficient and unified multimodal interactive coding framework that enables seamless edge--cloud communication between humans and AI agents.  
Unlike traditional pixel-based pipelines, UniMIC adopts a token-based paradigm that is natively compatible with LMMs, yielding substantial gains in both compression efficiency and downstream task performance.  
On the edge side, multimodal inputs---including images and text---are converted into discrete token sequences via modality-specific tokenizers.  
These tokens are entropy-coded and transmitted to the cloud, where a unified Transformer-based model directly processes them for diverse generation and understanding tasks, such as text-to-image synthesis, text-guided image editing, and VQA.  
After cloud-side processing, only the resulting tokens are entropy-coded and sent back to the edge, where they are reconstructed into final images or text outputs.

\figref{framework} provides an overview of the UniMIC architecture.  
In \secref{overview}, we describe the end-to-end workflow of UniMIC.  
We then introduce a lightweight Transformer-based entropy model in \secref{entropy}, designed to accurately estimate token probabilities and fully exploit redundancy across different transmission scenarios.  
Finally, \secref{protocol} details how UniMIC adapts to various multimodal tasks, demonstrating its flexibility and efficiency in real-world human--AI collaboration.

\subsection{Overview of the UniMIC Framework}
\label{sec:overview}
We refer to the human-operated device as the \emph{edge side} and the AI-agent host as the \emph{cloud side}.  
The edge side is assumed to have limited computational and memory resources, whereas the cloud side runs large-scale multimodal models, such as Transformer-based LMMs.

Depending on the task, the edge transmits different multimodal inputs.  
For T2I generation, only a text prompt $T_s$ is sent; for tasks such as text-guided image editing or VQA, both a text prompt $T_s$ and an image $I_s$ are required.  
We define the edge-side input as
\begin{equation}
    \mathcal{X}_s = \{ T_s, I_s \},
\end{equation}
where either element may be absent depending on the task.

These inputs are mapped into a \emph{unified token space} via modality-specific tokenizers:
\begin{align}
    \mathbf{t} &= \mathcal{E}_{\text{text}}(T_s) = \{ t_1, \ldots, t_N \}, \\
    \mathbf{u} &= \mathcal{E}_{\text{img}}(I_s) = \{ u_1, \ldots, u_M \},
\end{align}
The resulting text tokens $\mathbf{t}$ and image tokens $\mathbf{u}$ are concatenated as
\begin{equation}
    \mathbf{z} = [\mathbf{t}; \mathbf{u}],
\end{equation}
then entropy-encoded into a bitstream $B_{\text{up}}$ for \emph{edge-to-cloud} transmission (with text and image tokens coded separately; see \secref{entropy}).

On the cloud side, $B_{\text{up}}$ is decoded to recover
\begin{equation}
    \hat{\mathbf{z}} = [\hat{\mathbf{t}}; \hat{\mathbf{u}}],
\end{equation}
to which a task token $T_{\text{task}}$ is prepended, yielding
\begin{equation}
    \hat{\mathbf{z}}' = [T_{\text{task}}; \hat{\mathbf{z}}],
\end{equation}
the input to the \emph{Unified Transformer} $\mathcal{F}(\cdot)$.  
The model outputs a token sequence
\begin{equation}
    \mathbf{o} = \mathcal{F}(\hat{\mathbf{z}}'),
\end{equation}
which is entropy-encoded as $B_{\text{down}}$ for \emph{cloud-to-edge} transmission.

Upon receiving $B_{\text{down}}$, the edge decodes $\hat{\mathbf{o}}$ and reconstructs either an image $\hat{I}_t = \mathcal{D}_{\text{img}}(\hat{\mathbf{o}})$ or text $\hat{T}_t = \mathcal{D}_{\text{text}}(\hat{\mathbf{o}})$, depending on the task.

This token-centric pipeline avoids repeated pixel-level compression--decompression, preserves structural and semantic integrity, and minimizes transmission overhead while maintaining fidelity for both human viewing and downstream machine analysis.

\subsection{Entropy Modeling in Unified Token Space}
\label{sec:entropy}
Although UniMIC represents all multimodal inputs in a unified token space, the statistical properties of text tokens and image tokens differ significantly, necessitating modality-specific entropy coding strategies on both the edge and cloud sides.

\Paragraph{Text Tokens.}
Unlike conventional approaches that directly compress textual prompts, UniMIC first tokenizes text prompts into discrete tokens using a text tokenizer $\mathcal{E}_{\text{text}}(\cdot)$.  
The resulting token sequence $\mathbf{t}$ is then compressed losslessly using the Brotli~\cite{alakuijala2018brotli} algorithm, which is lightweight and well-suited for the typically small size of text token sequences:
\begin{equation}
    B_{\text{text}} = \text{Brotli}(\mathbf{t}).
\end{equation}

\begin{figure}[!t]
    \centering
 \includegraphics[width=1.0\linewidth]{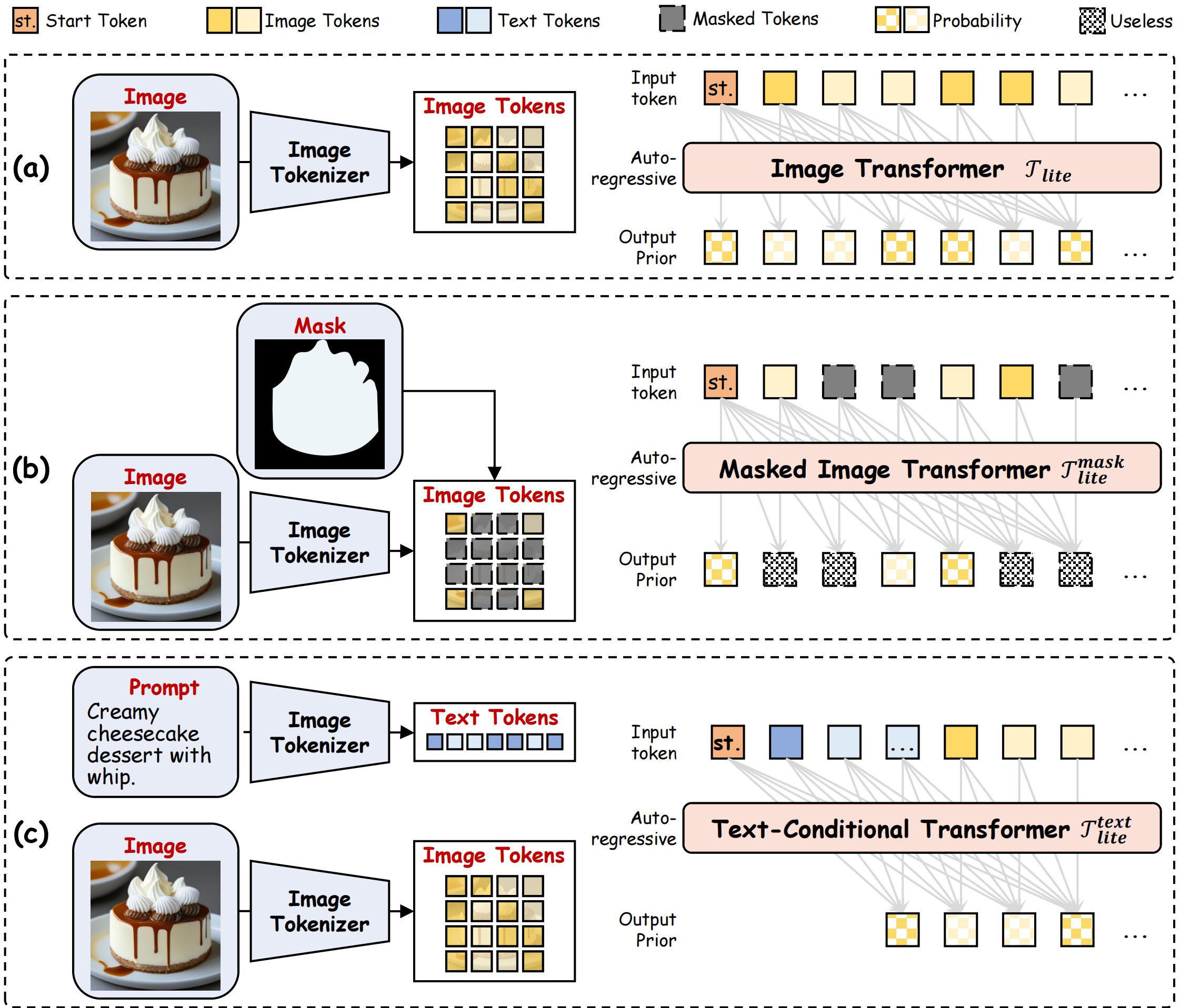}
 \caption{\textbf{Entropy modeling strategies for image tokens in UniMIC}, including (a) autoregressive, (b) masked-token, and (c) text-conditional mode.}
 \vspace{- 5mm}
    \label{fig:transformer_mode}
\end{figure}

\Paragraph{Image Tokens.}
Image tokens constitute the majority of the bitstream and exhibit much higher entropy, requiring specialized modeling to achieve efficient compression.  
To this end, we design a lightweight transformer-based entropy model, denoted as $\mathcal{T}_{\text{lite}}$, which estimates the probability distribution of image tokens to support arithmetic coding.  
Importantly, $\mathcal{T}_{\text{lite}}$ is designed to be sufficiently lightweight to be deployed on both the edge side and the cloud side, ensuring symmetric probability estimation and exact reconstruction.

Depending on the nature of the downstream task, we employ three variants of image-token entropy modeling, as illustrated in \figref{transformer_mode} :

\begin{itemize}
    \item \textbf{Autoregressive Mode $\mathcal{T}_{\text{lite}}$.}  
    For standard image token sequences, the joint probability is factorized autoregressively as:
    \begin{equation}
        p(\mathbf{u}) = \prod_{i=1}^{M} p(u_i \mid \mathbf{u}_{<i}; \theta_{\text{lite}}),
    \end{equation}
    where $\mathbf{u}$ is the image token sequence and $\theta_{\text{lite}}$ denotes the parameters of $\mathcal{T}_{\text{lite}}$.

  \item \textbf{Masked-Token Mode $\mathcal{T}^{\text{mask}}_{\text{lite}}$.}  
For tasks that require localized image editing, such as inpainting, only a subset of the image tokens needs to be updated.  
To efficiently handle these scenarios, we adopt a masked image transformer, $\mathcal{T}_{\text{lite}}^{\text{mask}}$, pre-trained using random masking strategies to learn the distribution of partially visible image token sequences.  
During training, we randomly select a subset of tokens in the original image token sequence $\mathbf{u} = \{ u_1, \ldots, u_M \}$ to be masked, producing a binary mask $\mathbf{m} = \{ m_1, \ldots, m_M \}$, where $m_i = 1$ denotes a masked position.  
These positions are replaced with a special mask token $\langle \text{MASK} \rangle$, yielding the masked sequence $\tilde{\mathbf{u}}$:
\begin{equation}
    \tilde{u}_i =
    \begin{cases}
        \langle \text{MASK} \rangle, & \text{if } m_i = 1, \\
        u_i, & \text{otherwise}.
    \end{cases}
\end{equation}

Importantly, although the masked tokens are not themselves targets for prediction, they remain in the context sequence as placeholders, providing positional and semantic cues to the transformer.  
Thus, the joint probability of all unmasked tokens is modeled as:
\begin{equation}
    p(\mathbf{u}_{\text{valid}} \mid \tilde{\mathbf{u}})
    = \prod_{i \in \mathcal{U}}
    p\bigl(u_i \mid \tilde{\mathbf{u}}_{<i}; \theta_{\text{lite}}^{\text{mask}} \bigr),
\end{equation}
where $\mathcal{U} = \{ i \mid m_i = 0 \}$ denotes the indices of unmasked tokens.
Accordingly, the entropy modeling loss is defined as:
\begin{equation}
    \mathcal{L}_{\text{entropy}} 
    = - \sum_{i \in \mathcal{U}}
    \log p\bigl(u_i \mid \tilde{\mathbf{u}}_{<i}; \theta_{\text{lite}}^{\text{mask}} \bigr).
\end{equation}
During encoding, only the unmasked tokens $\mathbf{u}_{\text{valid}}$ are entropy-coded and transmitted:
\begin{equation}
    \mathbf{u}_{\text{valid}} = \{ u_i \mid m_i = 0 \},
\end{equation}
On the decoder side, the binary mask $\mathbf{m}$ is reconstructed from the bitstream and used to re-insert mask tokens $\langle \text{MASK} \rangle$ at the masked positions, forming the reconstructed masked sequence $\mathbf{u}_{\text{mask}}$:
\begin{equation}
    \mathbf{u}_{\text{mask}}(i) =
    \begin{cases}
        u_i, & \text{if } m_i = 0, \\
        \langle \text{MASK} \rangle, & \text{if } m_i = 1.
    \end{cases}
\end{equation}
This masked sequence is then used as context for decoding the bitstream, ensuring that all unmasked tokens can be reconstructed precisely while avoiding the need to transmit unchanged image regions.  
This approach significantly improves compression efficiency by leveraging partial information while preserving spatial and semantic consistency.

\item \textbf{Text-Conditional Mode $\mathcal{T}^{\text{text}}_{\text{lite}}$.}  
In many multimodal applications, textual information is semantically correlated with visual content.  
A prime example is T2I generation, where the textual prompt $T_s$ serves as a direct description of the visual scene to be synthesized.    
During cloud-to-edge transmission, returning generated images from the cloud side to the edge side, both transmission directions contain textual information that can serve as powerful contextual priors for modeling the image token distribution.
To exploit these cross-modal relationships and reduce redundancy between the textual and visual modalities, 
we propose a \emph{text-conditional} entropy modeling strategy, where the probability of each image token $u_i$ is conditioned not only on its preceding image tokens $u_{<i}$, but also on the full text token sequence $\mathbf{t}$:
\begin{equation}
    p(\mathbf{u} \mid \mathbf{t})
    = \prod_{i=1}^{M} p\bigl( u_i \mid \mathbf{u}_{<i}, \mathbf{t}; \theta_{\text{lite}}^{\text{text}} \bigr),
\end{equation}
where $\theta_{\text{lite}}^{\text{text}}$ denotes the parameters of the text-conditional lightweight transformer.
This formulation allows the model to leverage semantic cues from the text tokens to more accurately predict the probabilities of image tokens, effectively reducing entropy and achieving better compression efficiency in multi-modal scenarios such as T2I generation.

\end{itemize}
% illustrates the architecture of the proposed entropy modeling schemes.

\begin{table}[!t]
\centering
\scriptsize
\caption{Task-specific token-based coding protocols in UniMIC. 
AR: Autoregressive Mode. 
Mask-T: Masked-Token Mode. 
Text-T: Text-Conditional Mode. 
Brotli: lossless compression for text tokens. 
EPAC: Equi-Probable Arithmetic Coding.}
\label{tab:task-protocols}
\begin{tabular}{l|c|c}
\hline
\textbf{Task} & \textbf{Edge-to-Cloud} & \textbf{Cloud-to-Edge} \\
\hline
T2I           & $\mathbf{t}$ (Brotli)             & $\hat{\mathbf{u}}$ (Text-T)           \\
Inpainting    & $\mathbf{u}_{\text{mask}}$ (Mask-T), $\mathbf{t}$ (Brotli), mask (EPAC) & $\hat{\mathbf{u}}_{\text{inpaint}}$ (Text-T) \\
Outpainting   & $\mathbf{u}$ (Text-T), $\mathbf{t}$ (Brotli) & $\hat{\mathbf{u}}_{\text{outpaint}}$ (Text-T) \\
VQA           & $\mathbf{u}$ (AR), $\mathbf{t}$ (Brotli) & $\hat{\mathbf{t}}_{\text{ans}}$ (Brotli) \\
\hline
\end{tabular}
\end{table}

\subsection{Task-Adaptive Transmission Protocols}
\label{sec:protocol}
In this section, we describe how UniMIC adapts its token-based coding framework to different multimodal human–AI collaborative tasks. Each task requires transmitting different combinations of tokenized modalities and employs corresponding entropy modeling strategies, yet all share the same unified token-based pipeline.
\tabref{task-protocols}  summarizes the token-based transmission protocols in UniMIC for different tasks, combining transmitted modalities with the corresponding entropy models used in both edge-to-cloud and cloud-to-edge directions.

\Paragraph{T2I Generation.}
The edge side transmits only text tokens $\mathbf{t}$, compressed losslessly using the Brotli~\cite{alakuijala2018brotli} algorithm.  
On the cloud side, the Unified Transformer takes $\mathbf{t}$ as input and generates the corresponding image token sequence $\hat{\mathbf{u}}$.  
Since the generated image tokens are strongly conditioned on the text prompt, the output sequence $\hat{\mathbf{u}}$ is entropy-coded using the text-conditional transformer $\mathcal{T}_{\text{lite}}^{\text{text}}$ before transmission back to the edge side.  
Upon receiving $\hat{\mathbf{u}}$, the edge side decodes and reconstructs the final synthesized image $I_t$ via the image detokenizer.

\Paragraph{Text-Guided Image Inpainting.}
Only the image tokens corresponding to regions that do not require editing (\ie unmasked tokens with $m_i=0$), together with a binary mask $\mathbf{m}$ indicating the positions of masked regions ($m_i=1$), and the text tokens $\mathbf{t}$ describing the desired edits are transmitted on the edge side.  
The text tokens $\mathbf{t}$ are compressed using the Brotli~\cite{alakuijala2018brotli} algorithm.  
The unmasked image tokens are entropy-coded via the masked-token transformer $\mathcal{T}_{\text{lite}}^{\text{mask}}$, since their encoding context includes placeholder mask tokens for the masked regions.  
On the cloud side, the Unified Transformer processes these inputs and predicts the token sequence $\hat{\mathbf{u}}_{\text{inpaint}}$ corresponding to the masked regions.  
These predicted tokens $\hat{\mathbf{u}}_{\text{inpaint}}$ are entropy-coded using the using the text-conditional transformer $\mathcal{T}_{\text{lite}}^{\text{text}}$ before being transmitted back to the user on the edge side.
Finally, the user merges the received inpainted tokens into the original sequence according to the binary mask, reconstructing the complete inpainted image.

\Paragraph{Text-Guided Image Outpainting.}
Both text tokens $\mathbf{t}$ and full image tokens $\mathbf{u}$ are transmitted from the edge to the cloud.
The text tokens $\mathbf{t}$ are compressed losslessly with Brotli~\cite{alakuijala2018brotli}, while the image tokens $\mathbf{u}$ are entropy-coded using the text-conditional transformer $\mathcal{T}_{\text{lite}}^{\text{text}}$, which leverages semantic cues from $\mathbf{t}$ to improve compression efficiency.
The Unified Transformer then generates additional tokens $\hat{\mathbf{u}}_{\text{outpaint}}$ representing the extrapolated regions.
These new tokens $\hat{\mathbf{u}}_{\text{outpaint}}$ are likewise compressed using $\mathcal{T}_{\text{lite}}^{\text{text}}$ and transmitted back to the user.
Finally, the user merges the received tokens with the original image tokens to reconstruct the outpainted image.

\Paragraph{VQA.}
The user on the edge side transmits both text tokens $\mathbf{t}$ (representing the question) and image tokens $\mathbf{u}$. 
The text tokens $\mathbf{t}$ are compressed using Brotli~\cite{alakuijala2018brotli}, while the image tokens $\mathbf{u}$ are entropy-coded via the autoregressive model $\mathcal{T}_{\text{lite}}$.  
On the cloud side, the Unified Transformer jointly processes these multimodal tokens and generates answer text tokens $\hat{\mathbf{t}}_{\text{ans}}$.  
The answer tokens are compressed with Brotli~\cite{alakuijala2018brotli} for efficient transmission.  
Once received, the user decodes the tokens to reconstruct the natural-language answer $T_t$.

\section{Experiments}
\label{sec:experiments}
We conduct extensive experiments to comprehensively validate the effectiveness of the proposed UniMIC framework across four representative multimodal human–AI collaborative tasks: T2I generation, text-guided image inpainting, text-guided image outpainting, and VQA.
Although these tasks share the same unified token-based communication infrastructure, they differ in terms of token compositions and entropy coding strategies, offering a diverse evaluation setting for our method.

%%%%%%%%%%%%%%%%%%%%%%%%%%%%%%%%%%%%%%%%%%%%%%%%%%%%%%%%%%%%%%%

\subsection{Implementation Details}

\Paragraph{Tokenizers and Unified Transformer Backbone.}
UniMIC is designed as a modular framework that can interface with unified multimodal transformer models capable of processing tokenized text and image inputs. 
In our experiments, we instantiate the backbone using the pretrained Show-o model~\cite{zhang2024show} for empirical validation, as it provides a strong off-the-shelf autoregressive transformer supporting both text-to-image generation and understanding. 
Text inputs are tokenized via a GPT-style Byte Pair Encoding (BPE) tokenizer~\cite{gpt2}, while images are encoded using MagViT-v2~\cite{magvit-v2}, which maps latent visual features into discrete tokens via an 8,192-entry codebook and Lookup-Free Quantization.
While Show-o~\cite{zhang2024show}  is adopted here for validation, the framework is not limited to this specific choice.
It can be integrated with other unified multimodal architectures with minimal adaptation, typically requiring retraining of the entropy modeling component to match the token distributions of a new backbone.

\Paragraph{Entropy Models and Training.}  
A key innovation of UniMIC lies in its entropy modeling strategy, which enables efficient compression of multimodal token sequences while preserving semantic fidelity.  
Text tokens are directly compressed using the Brotli~\cite{alakuijala2018brotli} lossless algorithm, while image tokens are handled by a lightweight autoregressive Transformer entropy estimator (0.6B parameters) combined with arithmetic coding, making it practical for deployment on both edge and cloud sides.  
To support diverse transmission scenarios, we employ three variants of this model: a standard Image Transformer $\mathcal{T}_{\text{lite}}$ for generic sequences, a Masked Image Transformer $\mathcal{T}^{\text{mask}}_{\text{lite}}$ for localized editing, and a Text-Conditional Image Transformer $\mathcal{T}^{\text{text}}_{\text{lite}}$ for text-guided generation.  
All models are implemented in PyTorch and trained on two NVIDIA A100 GPUs using a two-stage strategy:  
(1) \textbf{Pretraining} on large-scale datasets, where $\mathcal{T}_{\text{lite}}$ and $\mathcal{T}^{\text{mask}}_{\text{lite}}$ are trained on ImageNet-2012~\cite{deng2009imagenet} (with random masking for the masked variant), and $\mathcal{T}^{\text{text}}_{\text{lite}}$ is pretrained on MS COCO~\cite{lin2014microsoft} and CC3M~\cite{sharma2018conceptual} to capture text–image dependencies;  
(2) \textbf{Domain adaptation} to the Show-o token distribution via the synthetic \textit{Show-COCO} dataset, where images are re-tokenized by Show-o using their captions. The pretrained $\mathcal{T}^{\text{text}}_{\text{lite}}$ is fine-tuned on this dataset to yield $\mathcal{\hat{T}}^{\text{text}}_{\text{lite}}$, ensuring the entropy model aligns with the actual token statistics in cloud-to-edge transmission.

%%%%%%%%%%%%%%%%%%%%%%%%%%%%%%%%%%%%%%%%%%%%%%%%%%%
\begin{table}[!t]
\centering
\scriptsize
\caption{Quantitative comparisons on the T2I task.
The best results are highlighted in \textbf{bold} and the second-best results are \underline{underlined}.
``Lossless'' indicates perfect reconstruction, corresponding to infinite PSNR.}
\vspace{-2 mm}
\label{tab:t2i_new}
\begin{tabular}{lccccc}
\toprule
\textbf{Stage} & \multicolumn{3}{c}{\textbf{Cloud-to-Edge}} & \multicolumn{2}{c}{\textbf{Overall Process}} \\
\cmidrule(lr){2-4} \cmidrule(l){5-6}
\textbf{Metric} & BPP $\downarrow$ & PSNR $\uparrow$ & LPIPS $\downarrow$ & FID $\downarrow$ & CLIP-T $\uparrow$ \\
\midrule
\textbf{BPG~\cite{bellard2015bpg}}          & 0.0414 & 23.13 & 0.354 & 176.74 & 0.286 \\
\textbf{VVC~\cite{bross2021overview}}        & \underline{0.0337} & 23.12 & 0.343 & 180.19 & 0.286 \\
\textbf{MS-ILLM~\cite{ms-illm}}              & 0.0647 & \underline{25.76} & \underline{0.075} & 85.07 & 0.310 \\
\textbf{VQ-Kmeans~\cite{mao2024dcc}}         & 0.0431 & 22.03 & 0.102 & 82.53 & \underline{0.314} \\
\textbf{DiffEIC~\cite{diffeic}}              & 0.0587 & 20.13 & 0.179 & \underline{81.95} & 0.308 \\
\rowcolor{mycolor_blue} \textbf{UniMIC}             & \cellcolor{mycolor_blue}\textbf{0.0296} & \cellcolor{mycolor_blue}\textbf{lossless} & \cellcolor{mycolor_blue}\textbf{0} & \cellcolor{mycolor_blue}\textbf{80.61} & \cellcolor{mycolor_blue}\textbf{0.315} \\
\bottomrule
\vspace{-8 mm}
\end{tabular}
\end{table}
%%%%%%%%%%%%%%%%%%%%%%%%%%%%%%%%%%%%%%%%%%%%%%%

\subsection{Evaluation Details}
\label{sec:evaluationdetails}
\Paragraph{Baseline Compression Methods.}
We benchmark UniMIC against a set of representative codecs spanning both conventional and generative paradigms. 
For conventional compression, we adopt BPG~\cite{bellard2015bpg} (an HEVC-Intra-based image codec) and the latest VVC standard~\cite{bross2021overview}.  
For learned generative approaches, we include three representative methods: MS-ILLM~\cite{ms-illm}, a GAN-based codec; VQ-Kmeans~\cite{mao2024dcc}, based on VQGAN representations; and DiffEIC~\cite{diffeic}, which leverages diffusion models for image compression. 
All baselines are evaluated using publicly available pretrained weights and are configured to achieve bitrate levels comparable to UniMIC, ensuring a fair and consistent comparison.

\Paragraph{Common Compression Metrics.}
We adopt standard metrics to assess rate–distortion performance. 
Compression efficiency is measured in bits per pixel (bpp). 
Reconstruction quality is evaluated using Peak Signal-to-Noise Ratio (PSNR) for pixel-level fidelity and Learned Perceptual Image Patch Similarity (LPIPS) for perceptual similarity.
These metrics are applied consistently across all tasks to provide an objective assessment of the underlying compression schemes.
%%%%%%%%%%%%%%%%%%%%%%%%%%%%%%%%%%%%%%%%%%%%%%%
\begin{figure*}[!t]
    \includegraphics[width=1.0\linewidth]{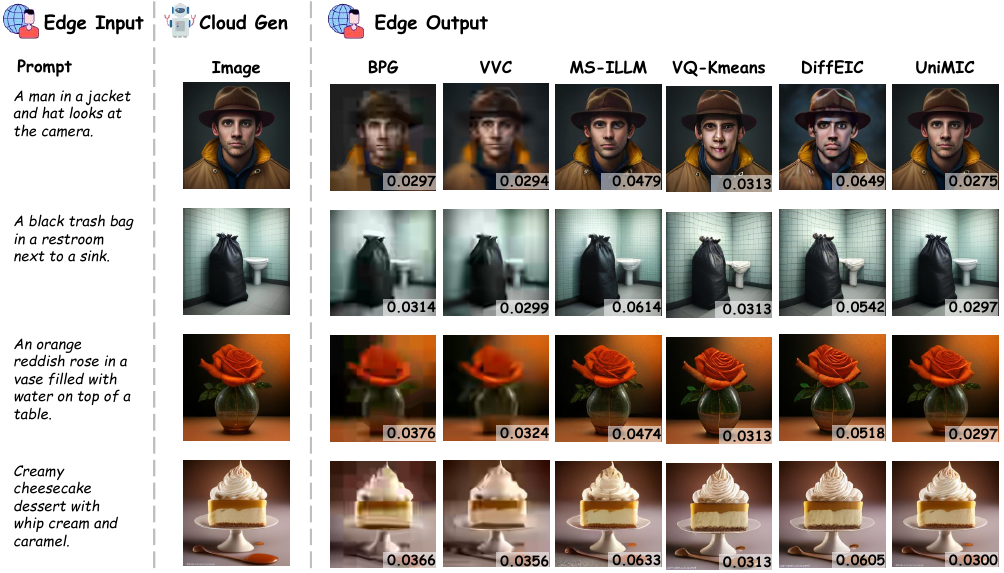}
\caption{\textbf{Qualitative comparison of T2I generation.} 
Compared with traditional codecs and recent generative baselines, UniMIC preserves semantic details and visual fidelity while achieving lower bitrates (bpp values shown under each result).}
    \label{fig:t2i_results}
\vspace{-5 mm}
\end{figure*}
%%%%%%%%%%%%%%%%%%%%%%%%%%%%%%%%%%%%%%%%%%%%%%%%%%%%%%%%%%%%%%%%%

\subsection{Compression Evaluation for Text-to-Image Generation}
\label{sec:comparison_generation}
We evaluate UniMIC on the T2I generation task, focusing on (i) the efficiency of image token compression during the cloud-to-edge transmission stage and (ii) the overall impact on text-conditioned image generation quality.

%\Paragraph{Evaluation Protocol.}
\Paragraph{Evaluation Protocol.}
We randomly sample $1{,}000$ human-annotated text prompts from the MS COCO 2017 validation set~\cite{lin2014microsoft} as inputs to the T2I generation pipeline.  
In the UniMIC pipeline, the edge side transmits only the tokenized text instructions, which are compressed using Brotli~\cite{alakuijala2018brotli}.  
On the cloud side, the Unified Transformer synthesizes the corresponding image tokens, which are then entropy-coded and transmitted back to the edge for reconstruction.  
For comparison, we follow the baseline selection in \secref{evaluationdetails} and benchmark UniMIC against both conventional codecs (BPG~\cite{bellard2015bpg}, VVC~\cite{bross2021overview}) and generative codecs (MS-ILLM~\cite{ms-illm}, VQ-Kmeans~\cite{mao2024dcc}, DiffEIC~\cite{diffeic}).  
Unlike UniMIC, these codecs cannot operate directly on tokens. Instead, the cloud-generated image tokens must first be decoded into full-resolution images before pixel-level compression is applied, and the compressed bitstream is then transmitted back to the edge side.

\Paragraph{Task-Specific Metrics.}
In addition to the common compression metrics introduced in \secref{evaluationdetails} (bpp, PSNR, LPIPS), 
we report two task-specific measures for T2I generation. 
Fréchet Inception Distance (FID)~\cite{heusel2017gans} evaluates the perceptual realism of generated images by measuring distributional divergence from real images, 
while CLIP-T~\cite{hessel2021clipscore} quantifies semantic alignment between generated images and input text prompts.

\Paragraph{Results and Analysis.}
\tabref{t2i_new} reports quantitative results under comparable bitrate conditions.  
UniMIC achieves near-lossless reconstruction even at extremely low bitrates ($\sim$0.03 bpp), consistently outperforming all baselines in PSNR and LPIPS.  
This advantage stems from directly entropy-coding discrete image tokens, which avoids pixel-domain re-encoding and the quantization losses inherent to other codecs.  
In addition, UniMIC obtains the lowest FID and highest CLIP-T scores, indicating superior preservation of both perceptual realism and semantic alignment with input prompts.  
As shown in \figref{t2i_results}, traditional codecs exhibit blurring and blocking artifacts, while generative codecs often introduce semantic shifts or structural distortions.  
In contrast, UniMIC reconstructs images that remain visually indistinguishable from their original cloud-generated counterparts, delivering perceptually lossless quality while maintaining high compression efficiency.

%%%%%%%%%%%%%%%%%%%%%%%%%%%%%%%%%%%%%%%%%%%%%%%%%%%%%%%%%%%%%%%%%
\begin{figure*}[!t]
    \centering
    \includegraphics[width=1.0\linewidth]{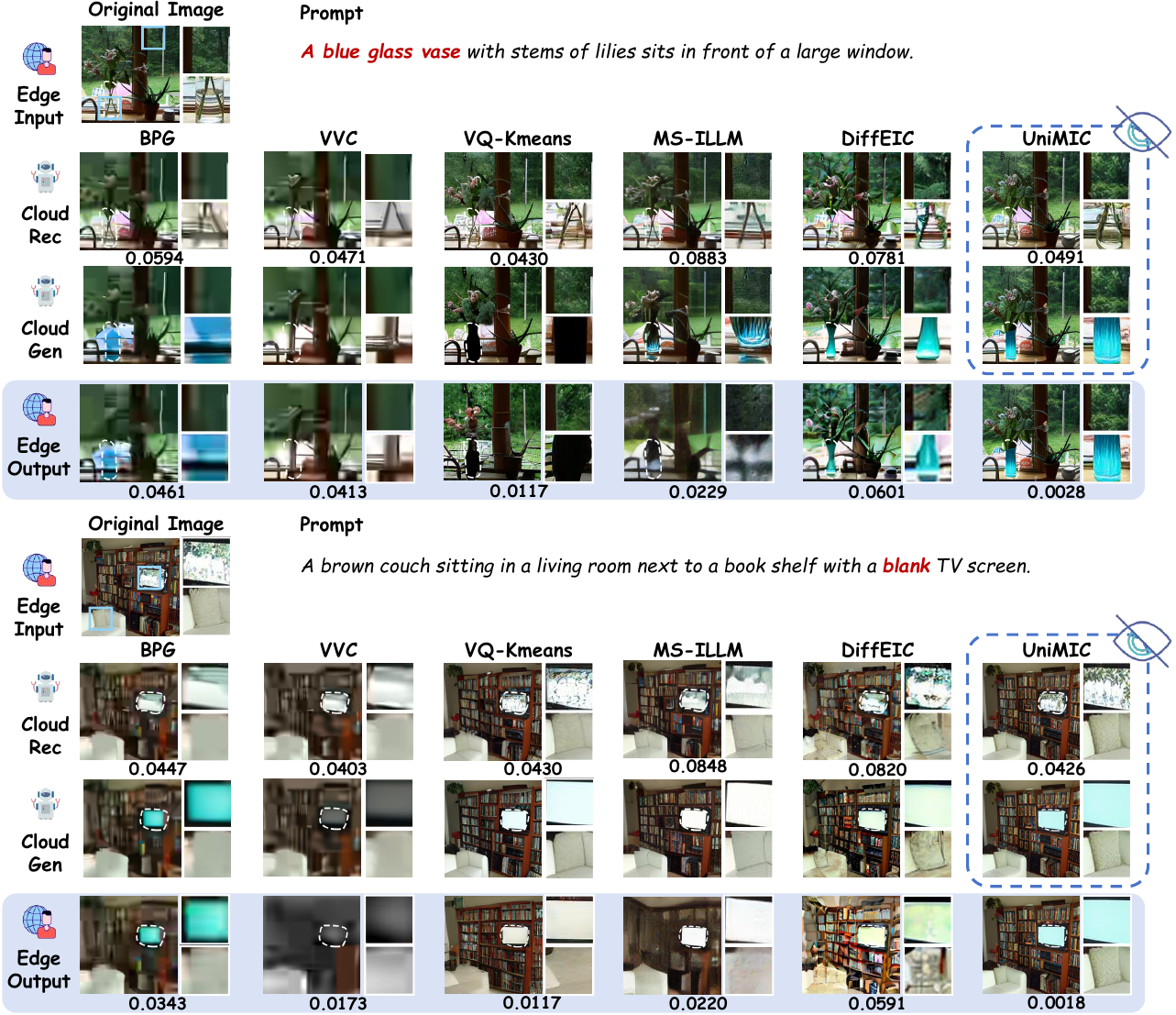}
   \caption{\textbf{Qualitative comparison of text-guided inpainting.} 
Baseline codecs suffer cumulative degradation: compression loss from Edge-to-Cloud is amplified during Cloud-to-Edge retransmission. 
In contrast, UniMIC incurs only a one-time tokenization loss, avoiding repeated degradation while preserving fidelity at ultra-low bitrates.
For clarity, note that the Cloud Rec and Cloud Gen stages do not exist in UniMIC’s actual pipeline; we reconstruct them here using original tokens and edited tokens solely to enable fair visual comparison with baselines.
}
\vspace{-5 mm}
\label{fig:inpainting_results}
\end{figure*}

\begin{table*}[!t]
\centering
\caption{Quantitative comparison results on text-guided image inpainting.
The best results are highlighted in \textbf{bold} and the second-best results are \underline{underlined}.
``Lossless'' indicates perfect reconstruction, corresponding to infinite PSNR.}
\label{tab:inpainting_results_table}
\begin{tabular}{l *{10}{c}}
\toprule
\textbf{Stage} & 
\multicolumn{3}{c}{\textbf{Edge-to-Cloud}} & 
\multicolumn{3}{c}{\textbf{Cloud-to-Edge}} & 
\multicolumn{4}{c}{\textbf{Overall Process}} \\
\cmidrule(lr){2-4} \cmidrule(lr){5-7} \cmidrule(l){8-11}
\textbf{Metric} & 
BPP $\downarrow$ & PSNR $\uparrow$ & LPIPS $\downarrow$ & 
BPP $\downarrow$ & PSNR $\uparrow$ & LPIPS $\downarrow$ & 
Total BPP $\downarrow$ & FID $\downarrow$ & CLIP-T $\uparrow$ & R-CLIP-I $\uparrow$ \\
\midrule
\textbf{BPG}      & 0.0514 & \underline{22.35} & 0.440 & 0.0381 & \underline{25.98} & \underline{0.245} & 0.0895 & 209.05 & 0.237 & 0.691 \\
\textbf{VVC}       & \underline{0.0370} & 21.88 & 0.481 & \underline{0.0160} & 20.52 & 0.486 & 0.0530 & 254.76 & 0.206 & 0.662 \\
\textbf{MS-ILLM}   & 0.0734 & \textbf{23.68} & \textbf{0.120} & 0.0230 & 19.24 & 0.381 & 0.0964 & 184.76 & 0.227 & 0.711 \\
\textbf{VQ-Kmeans} & 0.0391 & 19.85 & 0.149 & 0.0118 & 16.90 & 0.284 & \underline{0.0509} & \underline{95.51} & \underline{0.261} & \underline{0.779} \\
\textbf{DiffEIC}   & 0.0493 & 15.53 & 0.282 & 0.0306 & 12.71 & 0.534 & 0.0799 & 248.71 & 0.206 & 0.675 \\
\rowcolor{mycolor_blue}
\textbf{UniMIC} & 
\textbf{0.0369} & 20.42 & \underline{0.134} & 
\textbf{0.0063} & \textbf{lossless} & \textbf{0} & 
\textbf{0.0432} & \textbf{57.30} & \textbf{0.298} & \textbf{0.903} \\
\bottomrule
\end{tabular}
\vspace{-5mm}
\end{table*}

\subsection{Compression Evaluation for Text-Guided Image Inpainting}
\label{comparison_inpainting}
We evaluate UniMIC on the text-guided image inpainting task with two primary objectives: (i) assessing the compression efficiency of image tokens across both edge-to-cloud and cloud-to-edge transmissions, and 
(ii) evaluating its effectiveness in text-conditioned editing.

\Paragraph{Evaluation Protocol.}
We employ the MagicBrush dataset, which provides 535 raw images and 1,053 edited samples with semantically masked regions and corresponding text instructions~\cite{Zhang2023MagicBrush}, as inputs to the text-guided inpainting pipeline. 
In UniMIC, the edge side transmits only the unedited image tokens, along with the binary mask and text prompt.  
Conditioned on these inputs, the Unified Transformer generates tokens for the masked regions, which are entropy-coded and returned to the edge, where they are merged with the unedited tokens to reconstruct the final inpainted image.  
We also follow the baseline selection in \secref{evaluationdetails} and evaluate UniMIC against conventional codecs and generative codecs.
Since these codecs cannot process tokens directly, the full image must be compressed and transmitted to the cloud for decompression and editing. 
After editing, the result is re-compressed and sent back to the edge.

\Paragraph{Task-Specific Metrics.}
We adopt the general compression metrics defined in~\secref{evaluationdetails}, including bpp, PSNR, and LPIPS, to evaluate reconstruction fidelity across both transmission stages.  
We compare reconstruction quality at two stages: edge-to-cloud reconstruction is evaluated against the original input image, while cloud-to-edge reconstruction is evaluated against the generated edited image.
In UniMIC, the edge-to-cloud stage transmits only the tokens of unedited regions rather than the full image.  
To ensure fair evaluation, we reconstruct the complete image on the edge side from the full token sequence and report PSNR/LPIPS on this reconstruction, while the bpp is computed using only the transmitted unedited-region tokens.  
For the cloud-to-edge stage, we further assess the quality of the inpainted outputs by reporting FID~\cite{heusel2017gans} for perceptual realism, CLIP-T~\cite{hessel2021clipscore} for semantic alignment with text prompts, and R-CLIP-I~\cite{hessel2021clipscore} for structural consistency with the original unedited regions.

\Paragraph{Results and Analysis.}
\tabref{inpainting_results_table} reports the quantitative results on the text-guided image inpainting task under comparable bitrate conditions.
UniMIC achieves the lowest transmission cost in both directions (0.0369 bpp for edge-to-cloud and 0.0063 bpp for cloud-to-edge), while also attaining the best overall performance on semantic and perceptual metrics (FID = 57.30, CLIP-T = 0.298, R-CLIP-I = 0.903).
Although the reconstructed images in the edge-to-cloud stage inevitably show some degradation due to the tokenizer, this loss is fixed and does not propagate further, since token sequences are transmitted and compressed losslessly.
In contrast, baseline codecs must process pixel-level images: compression artifacts are introduced during edge-to-cloud transmission, which then propagate through the pipeline and significantly impair the quality of inpainting and final reconstructions.

\figref{inpainting_results} provides qualitative comparisons under similar bitrates.
\emph{For fair visualization, we additionally reconstruct the intermediate stages (Cloud Rec, Cloud Gen) for UniMIC by using the complete token set, even though in the actual pipeline the cloud side does not have access to all tokens and only the final edge-side output is observable.}
For baseline methods, these stages are naturally available since they operate on pixel-level images.
In all cases, the reported bitrates strictly correspond to the actual tokens transmitted in UniMIC and the actual compressed images in baselines.
The visual results further confirm the quantitative findings: traditional codecs (VVC~\cite{bross2021overview} and BPG~\cite{bellard2015bpg}) already suffer from severe blurring and blocking artifacts in the edge-to-cloud reconstruction, and these distortions are further amplified during the cloud-to-edge transmission, leading to heavily degraded final outputs.
Generative codecs (DiffEIC~\cite{diffeic}, MS-ILLM~\cite{ms-illm}, and VQ-Kmeans~\cite{mao2024dcc}) produce more natural details, but the artifacts and inconsistencies introduced in the first stage also propagate into the second, often causing semantic shifts or unintended changes in unedited regions, resulting in noticeable divergence from the original content.
In contrast, UniMIC avoids this cumulative degradation since tokens are transmitted losslessly after the one-time tokenizer step, ensuring that the inpainted content generated in the cloud is faithfully preserved in the final edge-side reconstruction.

\begin{figure*}[!t]
    \centering
    \includegraphics[width=0.95\linewidth]{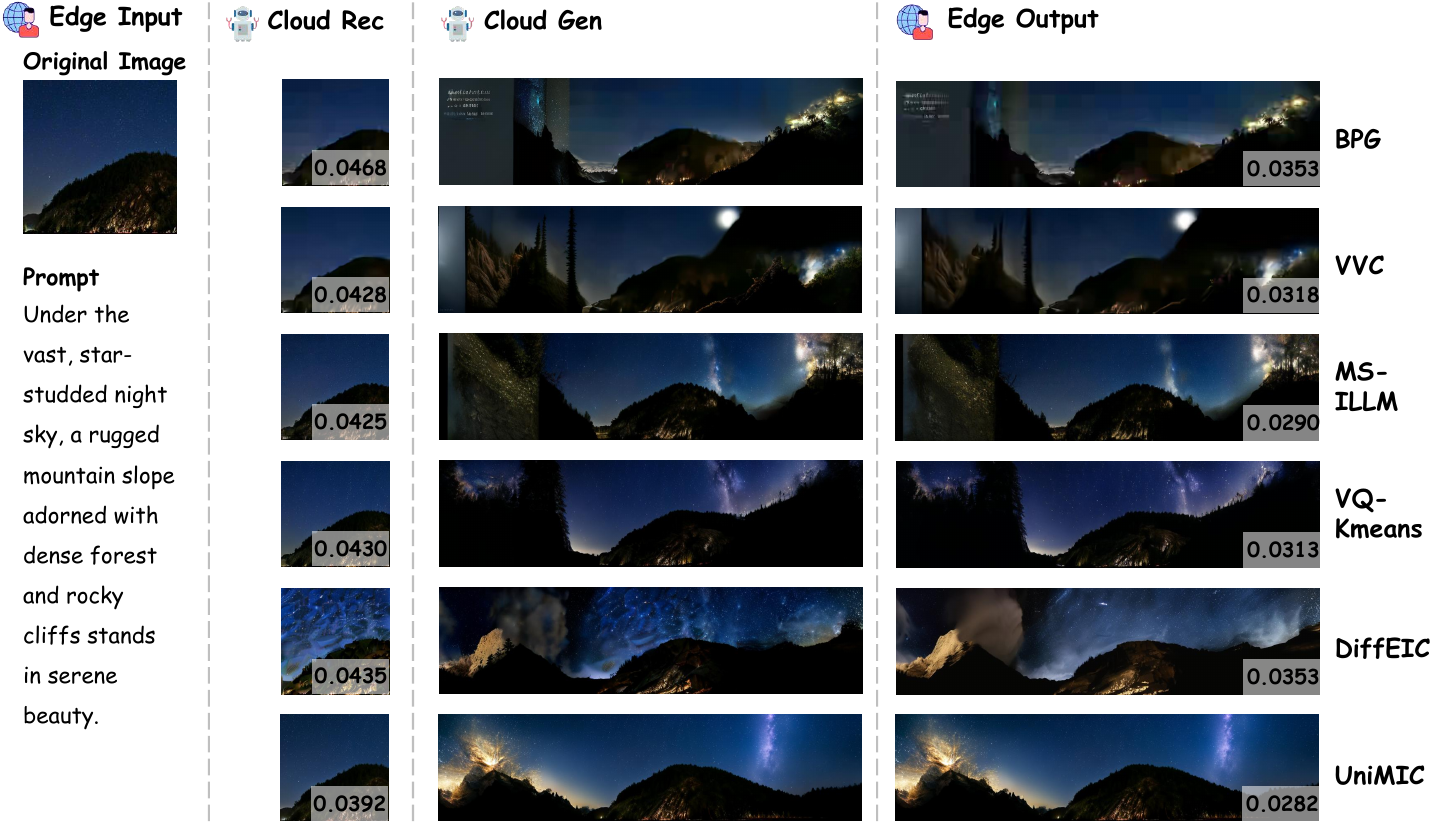}
    \vspace{-2mm}
    \caption{\textbf{Qualitative comparison of text-guided image outpainting.} 
Baseline methods accumulate distortion across transmission stages and often lose semantic consistency in the extended regions. 
UniMIC avoids repeated degradation and generates coherent outpainting at ultra-low bitrates.}
    \label{fig:outpainting_results}
    \vspace{-5 mm}
\end{figure*}

\begin{table*}[!t]
\centering
\caption{Quantitative Comparisons on the Text-Guided Image Outpainting Task.
The best results are highlighted in \textbf{bold} and the second-best results are \underline{underlined}.
``Lossless'' indicates perfect reconstruction, corresponding to infinite PSNR.}
\label{tab:outpainting}
\begin{tabular}{lccccccccc}
\toprule
\multirow{2}{*}{\textbf{Stage}} & \multicolumn{3}{c}{\textbf{Edge-to-Cloud}} & \multicolumn{3}{c}{\textbf{Cloud-to-Edge}} & \multicolumn{3}{c}{\textbf{Overall Process}} \\
\cmidrule(lr){2-4} \cmidrule(lr){5-7} \cmidrule(l){8-10}
& BPP$\downarrow$ & PSNR$\uparrow$ & LPIPS$\downarrow$ & BPP$\downarrow$ & PSNR$\uparrow$ & LPIPS$\downarrow$ & Total BPP$\downarrow$ & FID$\downarrow$ & CLIP-T$\uparrow$ \\
\midrule
\textbf{BPG\cite{bellard2015bpg}} & 0.0610 & 24.39 & 0.484 & \underline{0.0387} & 28.69 & 0.274 & 0.0997 & 213.36 & 0.232 \\
\textbf{VVC\cite{bross2021overview}} & 0.0576 & \textbf{24.84} & 0.524 & 0.0387 & \underline{29.69} & 0.184 & 0.0963 & 216.88 & 0.252 \\
\textbf{MS-ILLM\cite{ms-illm}} & 0.0622 & \underline{24.50} & \textbf{0.126} & 0.0396 & 27.60 & \underline{0.083} & 0.1018 & 56.41 & 0.280 \\
\textbf{VQ-Kmeans\cite{mao2024dcc}} & \underline{0.0431} & 20.12 & 0.139 & 0.0313 & 23.45 & 0.105 & 0.0744 & \underline{39.16} & \underline{0.290} \\
\textbf{DiffEIC\cite{diffeic}} & 0.0437 & 16.37 & 0.228 & 0.0401 & 19.69 & 0.187 & 0.0838 & 66.73 & 0.287 \\
\rowcolor{mycolor_blue} 
\textbf{UniMIC} & \cellcolor{mycolor_blue}\textbf{0.0416} & \cellcolor{mycolor_blue}22.07 & \cellcolor{mycolor_blue}\underline{0.127} & \cellcolor{mycolor_blue}\textbf{0.0273} & \cellcolor{mycolor_blue}\textbf{lossless} & \cellcolor{mycolor_blue}\textbf{0} & \cellcolor{mycolor_blue}\textbf{0.0689} & \cellcolor{mycolor_blue}\textbf{30.04} & \cellcolor{mycolor_blue}\textbf{0.291} \\
\bottomrule
\end{tabular}
\vspace{-3mm}
\end{table*}

\subsection{Compression Evaluation for Text-Guided Image Outpainting}
\label{comparison_outpainting}
We evaluate UniMIC on the text-guided image outpainting task, focusing on 
(i) the compression efficiency of image tokens during both edge-to-cloud and cloud-to-edge transmissions, and 
(ii) its effectiveness in generating high-quality extrapolated content conditioned on textual instructions.

\Paragraph{Evaluation Protocol.}
We construct a large-scale test set based on the Flickr Scenery dataset~\cite{lin2021infinity}, from which $1{,}000$ high-resolution landscape images are randomly selected.  
To enable text-guided outpainting, we employ Qwen-VL~\cite{wang2024qwen2} to generate descriptions for extrapolated regions, forming paired inputs of original images and text prompts.  
Following the baseline selection in \secref{evaluationdetails}, we compare UniMIC against conventional codecs (BPG, VVC) and generative codecs (MS-ILLM~\cite{ms-illm}, VQ-Kmeans~\cite{mao2024dcc}, DiffEIC~\cite{diffeic}) under a unified two-stage transmission pipeline comprising edge-to-cloud and cloud-to-edge communication.  
In UniMIC, the edge side compresses and transmits image tokens—rather than pixel-level images—to the cloud, where the Unified Transformer performs outpainting directly in the token space conditioned on the received tokens and text prompt.  
The generated tokens for the extrapolated regions are then entropy-coded and transmitted back to the edge, where they are seamlessly merged with the original tokens to reconstruct the final image.  
In contrast, baseline codecs cannot operate on tokens. The edge-side image must first be compressed and sent to the cloud, where it is decompressed, outpainted based on the input prompt, and the resulting full-resolution image is re-compressed for cloud-to-edge transmission.  
We report reconstruction quality at both stages: edge-to-cloud reconstruction is measured against the original input image, while cloud-to-edge reconstruction is evaluated against the generated outpainted image.  

\Paragraph{Task-Specific Metrics.}
We adopt the same evaluation metrics as in the T2I generation task (cf. \secref{comparison_generation}), including bpp, PSNR, LPIPS, FID, and CLIP-T.
While bpp, PSNR, and LPIPS jointly reflect the reconstruction fidelity during both transmission stages, FID specifically quantifies the semantic and perceptual quality of the extrapolated content conditioned on textual prompts.
CLIP-T further evaluates the alignment between the extrapolated image regions and the guiding text, measuring whether the generated visual content faithfully follows the textual instructions.

\Paragraph{Results and Analysis.}  
\tabref{outpainting} reports the quantitative results for text-guided image outpainting under comparable bitrate settings.  
In the edge-to-cloud stage, UniMIC achieves the lowest bitrate while retaining competitive perceptual quality and distortion performance.  
Unlike conventional codecs that must decode the original image before compression, UniMIC directly operates on discrete tokens without reconstructing pixel-level images.  
For fairness, we employ an auxiliary image detokenizer to visualize cloud-side reconstructions; the slight degradation observed arises from the inherent lossy tokenizer–detokenizer process, rather than the token-level compression itself.  
In the cloud-to-edge stage, UniMIC delivers lossless reconstruction even below $0.03$ bpp, its discrete token design allows transmission of only the newly generated tokens for extrapolated regions, which are seamlessly merged with unedited tokens on the edge side, substantially reducing bandwidth while preserving semantic coherence.  
As a result, UniMIC also achieves the lowest FID score, demonstrating superior perceptual realism of the extrapolated content.  

Beyond these quantitative improvements, qualitative results further highlight UniMIC’s advantages.
\figref{outpainting_results} compares different methods across edge-to-cloud reconstruction, cloud-side outpainting, and cloud-to-edge reconstruction.  
Baseline codecs introduce compression artifacts during edge-to-cloud transmission, which degrade the cloud input and subsequently impair the extrapolated content despite identical text prompts.  
Pixel-level re-compression of the outpainted image further accumulates distortions, producing visible inconsistencies in the final reconstructions.  
In contrast, UniMIC maintains structural fidelity and stylistic consistency, yielding natural and coherent extensions aligned with both the original image and the guiding prompt.  
Among conventional codecs, VVC~\cite{bross2021overview} and BPG~\cite{bellard2015bpg} often cause blurring, color shifts, and discontinuities between original and extrapolated regions, while generative methods such as DiffEIC~\cite{diffeic}, MS-ILLM~\cite{ms-illm}, and VQ-Kmeans~\cite{mao2024dcc} improve perceptual quality but still suffer from semantic drifts and texture inconsistencies, leading to stylistic deviations from the source image.

% 定量评估

% 

 \begin{figure*}[!t]
    \centering
    \includegraphics[width=1.0\linewidth]{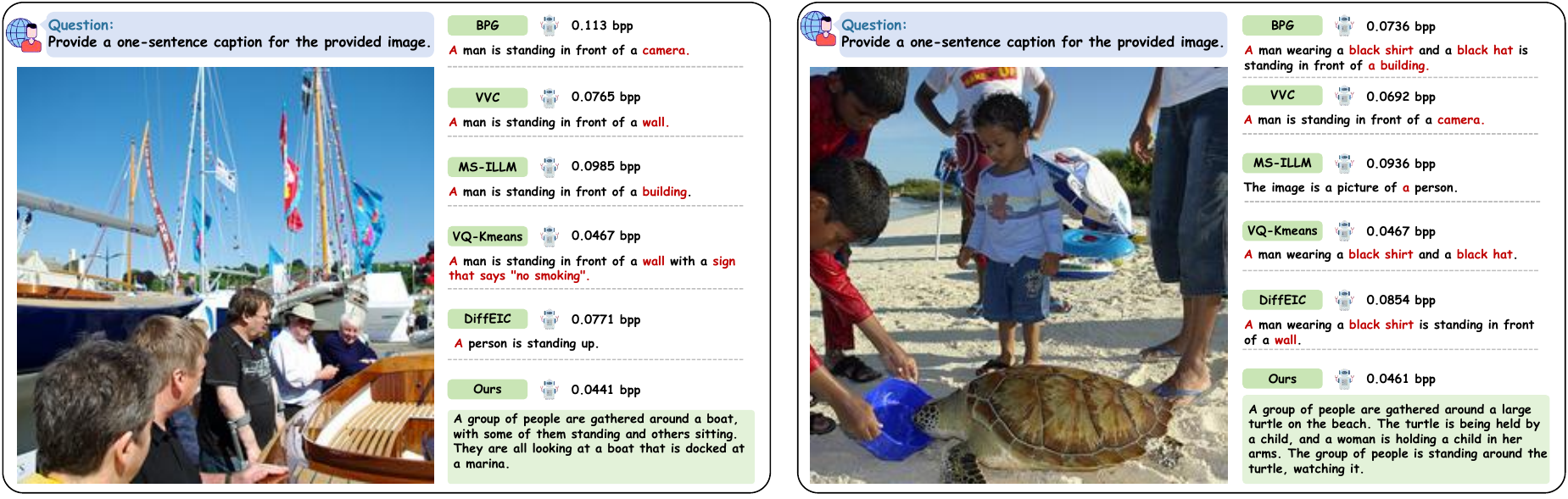}
    \vspace{-2mm}
    \caption{\textbf{Qualitative comparison of VQA capability on examples from the Flickr30k dataset~\cite{young2014image}.}  We show two representative cases with input questions and the corresponding answers generated by different methods. Baseline codecs (BPG, VVC, MS-ILLM, VQ-Kmeans, DiffEIC) either produce generic or semantically incorrect responses, whereas UniMIC generates contextually rich and accurate captions that align closely with the visual content.
    }
    \label{fig:vqa_results}
    \vspace{-3mm}
\end{figure*}
\begin{table*}[!t]
\vspace{-1mm}
\centering
\caption{Quantitative comparison results across multiple VQA datasets. 
The best results are highlighted in \textbf{bold} and the second-best results are \underline{underlined}.
``Lossless'' indicates perfect reconstruction, corresponding to infinite PSNR.
}
\label{tab:vqa}
\begin{tabular}{lcccccccccc}  % 11 columns total (was 12)
\toprule
\textbf{Stage} &
\multicolumn{2}{c}{\textbf{POPE~\cite{Li-hallucination-2023}}} &
\multicolumn{2}{c}{\textbf{GQA~\cite{hudson2019gqa}}} &
\multicolumn{4}{c}{\textbf{Flickr-30k~\cite{young2014image}}} &  % Now 4 metrics
\multicolumn{2}{c}{\textbf{Vizwiz-val~\cite{gurari2018vizwiz}}} \\
\cmidrule(lr){2-3} \cmidrule(lr){4-5} \cmidrule(lr){6-9} \cmidrule(lr){10-11}  % Adjusted: 6-9 for Flickr, 10-11 for Vizwiz
& BPP$\downarrow$ & ACC$\uparrow$
& BPP$\downarrow$ & EM$\uparrow$
& BPP$\downarrow$ & BLEU$\uparrow$ & CIDEr$\uparrow$ & ROUGE-L$\uparrow$  % Removed METEOR
& BPP$\downarrow$ & EM$\uparrow$ \\
\midrule
\textbf{BPG~\cite{bellard2015bpg}}      & 0.0565 & 0.5290 & 0.0569 & 0.3357 & 0.0535 & \underline{0.1072} & \underline{0.0193} & \underline{0.2232} & \underline{0.0470} & \underline{0.0156} \\
\textbf{VVC~\cite{bross2021overview}}    & \underline{0.0465} & \underline{0.5348} & 0.0529 & 0.3318 & 0.0500 & 0.0968 & 0.0166 & 0.2142 & 0.0429 & 0.0125 \\
\textbf{MS-ILLM~\cite{ms-illm}}          & 0.0752 & 0.5267 & 0.0754 & 0.3312 & 0.0793 & 0.0969 & 0.0167 & 0.2145 & 0.0578 & 0.0107 \\
\textbf{VQ-Kmeans~\cite{mao2024dcc}}     & 0.0470 & 0.5312 & \underline{0.0470} & 0.3348 & \underline{0.0470} & 0.1022 & 0.0176 & 0.2203 & 0.0470 & 0.0128 \\
\textbf{DiffEIC~\cite{diffeic}}          & 0.0740 & 0.5309 & 0.0747 & 0.0339 & 0.0770 & 0.1010 & 0.0177 & 0.2189 & 0.0620 & 0.0110 \\
\rowcolor{mycolor_blue}
\textbf{UniMIC}                            & \textbf{0.0424} & \textbf{0.7710} & \textbf{0.0422} & \textbf{0.4915} & \textbf{0.0429} & \textbf{0.2807} & \textbf{0.3398} & \textbf{0.4200} & \textbf{0.0394} & \textbf{0.0674} \\
\bottomrule
\end{tabular}
\vspace{-5mm}
\end{table*}

\subsection{Compression Evaluation for Visual Question Answering}
We evaluate UniMIC on the VQA task, focusing on the efficiency of image token compression during the edge-to-cloud transmission stage and the overall performance of answer generation.

\Paragraph{Evaluation Protocol.}  
We evaluate UniMIC on the VQA task using four benchmark datasets:  
\textbf{POPE}~\cite{Li-hallucination-2023}, built on MS COCO with 9,000 questions requiring precise object localization alongside adversarial distractors;  
\textbf{GQA}~\cite{hudson2019gqa}, comprising 20M reasoning questions paired with real-world images annotated with detailed scene graphs;  
\textbf{Flickr30k}~\cite{young2014image}, containing 31,783 images with descriptive captioning tasks; and  
\textbf{Vizwiz-val}~\cite{gurari2018vizwiz}, featuring 4,319 real-world image–question pairs collected from blind users, covering recognition, reading, and reasoning tasks.  
In UniMIC, the edge compresses and sends image and question tokens to the cloud, where VQA is performed directly in the token space.
The cloud then returns entropy-coded answer tokens, which are decoded on the edge into natural-language responses.
In contrast, baseline codecs operate at the pixel level: the original image must first be compressed and transmitted to the cloud, decompressed for VQA, and the resulting answer subsequently transmitted back.

\Paragraph{Task-Specific Metrics.}  
In line with the previous tasks, we report the edge-to-cloud image bpp to measure compression efficiency.  
For task performance, different metrics are adopted depending on the dataset and the nature of the text responses:  

\begin{itemize}
    \item \textbf{POPE~\cite{Li-hallucination-2023}:} Accuracy is used to assess the model’s ability to correctly recognize objects and their spatial relationships in the presence of distractors.  
    Higher accuracy reflects stronger fine-grained visual grounding.  

    \item \textbf{GQA~\cite{hudson2019gqa}:} We use Exact Match (EM) to measure the proportion of generated answers that exactly match the reference answers, where higher EM indicates stronger multimodal reasoning and semantic faithfulness.  

    \item \textbf{Flickr30k~\cite{young2014image}:} Captioning performance is evaluated with BLEU, CIDEr,and ROUGE-L.  
    Higher scores reflect better semantic alignment and descriptive quality relative to human-written captions.  

    \item \textbf{Vizwiz-val~\cite{gurari2018vizwiz}:} Similar to GQA~\cite{hudson2019gqa}, EM is employed to evaluate answer quality on this real-world dataset, where higher scores indicate stronger robustness to diverse and noisy visual–linguistic inputs.  
\end{itemize}

\begin{table*}[!t]
\centering
\caption{Ablation Study Results of Transformer-Based Entropy Models. 
Results for our final scheme are shown with a blue background.
The best results are highlighted in \textbf{bold} and the second-best results are \underline{underlined}.
}
\label{tab:image_transformer}
\renewcommand{\arraystretch}{1.2} % Increase row height for better readability
% 在 tabular 中使用 \bfseries 加粗表头
\begin{tabular}{l *{6}{c}}
\toprule
\multirow{2}{*}{\bfseries Methods} & \bfseries T2I & \multicolumn{2}{c}{\bfseries Inpainting} & \multicolumn{2}{c}{\bfseries Outpainting} & \bfseries VQA \\
\cmidrule(lr){2-2} \cmidrule(lr){3-4} \cmidrule(lr){5-6} \cmidrule(l){7-7}
& \bfseries Cloud-to-Edge & \bfseries Edge-to-Cloud & \bfseries Cloud-to-Edge & \bfseries Edge-to-Cloud & \bfseries Cloud-to-Edge & \bfseries Edge-to-Cloud \\ 
\midrule
\bfseries w/o Transformer & 0.0509 & 0.0429 & 0.0081 & 0.0509 & 0.0382 & 0.0509 \\
\bfseries $\mathcal{T}_{\text{lite}}$ & 0.0347 & 0.0385 & 0.0064 & \underline{0.0423} & \underline{0.0274} & \cellcolor{mycolor_blue}\textbf{0.0424}\\
\bfseries $\mathcal{T}^{\text{mask}}_{\text{lite}}$ & 0.0368 & \cellcolor{mycolor_blue}\textbf{0.0369} & 0.0065 & 0.0432 & 0.0276  & 0.0438 \\
\bfseries $\mathcal{T}^{\text{text}}_{\text{lite}}$ & \underline{0.0336} & \underline{0.0384} & \underline{0.0063} & \cellcolor{mycolor_blue}\textbf{0.0416} & \cellcolor{mycolor_blue}\textbf{0.0273} & \underline{0.0425} \\
\bfseries $\mathcal{\hat{T}}^{\text{text}}_{\text{lite}}$ & \cellcolor{mycolor_blue}\textbf{0.0296} & 0.0398 & \cellcolor{mycolor_blue}\textbf{0.0062} & 0.0432 & 0.0277 & 0.0440 \\
\bottomrule
\vspace{-5 mm}
\end{tabular}
\end{table*}

\begin{table}[!t]
\centering
\caption{Ablation study of text compression methods across different multimodal tasks.
The best results are highlighted in \textbf{bold}.
}
\label{tab:text_bpp}
\resizebox{0.95\linewidth}{!}{
\begin{tabular}{l c c c}
\toprule
\textbf{Task} & \textbf{Text Type} & \textbf{Brotli (CR$\downarrow$)} & \textbf{Ours (CR$\downarrow$)} \\
\midrule
\textbf{T2I}      & Prompt   & 0.905 & \textbf{0.745} \\
\textbf{Inpainting}  & Prompt   & 0.824 & \textbf{0.656} \\
\textbf{Outpainting} & Prompt   & 0.728 & \textbf{0.548} \\
\textbf{VQA}     & Question & 0.733 & \textbf{0.632} \\
\textbf{VQA}     & Answer   & 0.843 & \textbf{0.691} \\
\bottomrule
\end{tabular}
}
\vspace{-5 mm}
\end{table}

\Paragraph{Results and Analysis.}  
\tabref{vqa} reports quantitative comparisons across all datasets under comparable bitrate settings, while \figref{vqa_results} illustrates representative qualitative results.  
Baseline codecs significantly impair the semantic understanding of the cloud model after decompression, leading to degraded reasoning and inaccurate answers to textual queries.  
In contrast, UniMIC preserves the token-level information required by the cloud-based model, maintaining robust performance.  
Notably, UniMIC consistently achieves higher task-specific scores at substantially lower bitrates ($\sim$0.04 bpp), demonstrating that operating directly on discrete tokens eliminates the information loss caused by pixel-level compression and ensures stable end-to-end VQA performance.  
These results further validate the generality of UniMIC, showing that its token-based interactive coding framework extends beyond generation and editing tasks to support complex multimodal reasoning in real-world scenarios.

\subsection{Ablation Studies}
We conduct ablation experiments to quantify the contribution of key components in UniMIC. Specifically, we first evaluate different Transformer-based entropy modeling strategies and then investigate the effect of text tokenization on transmission efficiency.

\Paragraph{Impact of Transformer-Based Entropy Models.}
We conduct the ablation study in two stages to validate the effectiveness of Transformer-based entropy models in UniMIC.
First, compared to the baseline without Transformer modeling, all variants achieve clear bitrate reductions, \eg lowering T2I (Cloud-to-Edge) from $0.0509$ to $0.0296$, confirming the advantage of probabilistic modeling over a uniform token distribution.
Second, we evaluate the three proposed variants—Image Transformer, Masked Image Transformer, and Text-Conditional Transformer—under their respective task settings. Each shows superior performance in its intended scenario: the Text-Conditional model excels in text-to-image and outpainting, the Masked model is most effective for inpainting (Edge-to-Cloud), and the Image Transformer performs best on VQA. These results verify that our task-specific designs provide consistent gains across diverse transmission directions.
Finally, domain adaptation of the Text-Conditional Transformer on the Show-COCO dataset further reduces the bitrate for T2I (from $0.0336$ bpp to $0.0296$ bpp, an $11.9\%$ improvement), highlighting the importance of adapting entropy models to the token distributions of generative models.

\Paragraph{Effect of Text Tokenization on Compression Efficiency.}
UniMIC transmits textual inputs by first converting them into discrete tokens and then applying entropy coding, rather than compressing raw text directly.
To assess the effectiveness of this design, we compare UniMIC with a baseline where Brotli~\cite{alakuijala2018brotli} is directly applied to raw text sequences without tokenization.
We evaluate text compression efficiency using Compression Ratio (CR) across different multimodal tasks.
As shown in \tabref{text_bpp}, UniMIC achieves consistently better compression than the Brotli baseline, with improvements ranging from $13.8\%$ to $24.7\%$ across tasks.
These results validate that tokenizing text before entropy coding yields more compact representations and thus more efficient transmission in multimodal communication pipelines.

\section{Conclusions}
\label{sec:conclusions}
In this paper, we presented UniMIC, a unified token-based multimodal interactive coding framework for next-generation human–AI collaboration. By replacing conventional pixel- or text-based transmission with compact tokenized representations, UniMIC enables efficient bidirectional communication while preserving structural alignment with LMMs. To further enhance efficiency, we designed lightweight Transformer-based entropy models with scenario-specific strategies that effectively reduce redundancy and adapt to diverse tasks. Extensive experiments across T2I generation, text-guided inpainting, outpainting, and VQA validate the robustness of UniMIC, demonstrating substantial bitrate savings and stable performance even under ultra-low bitrate conditions. 
Beyond compression efficiency, UniMIC establishes a paradigm shift from pixel-centric coding to token-based communication protocols, paving the way for future AI-native multimedia transmission systems.
Nevertheless, UniMIC still relies on the quality and compatibility of upstream tokenizers, and adapting the entropy models to different LMM backbones may require lightweight retraining to mitigate domain gaps. We leave these aspects as promising directions for future work.
\renewcommand{\IEEEbibitemsep}{0pt plus 0.5pt}
\setlength{\itemsep}{-0.2em}
\setlength{\parsep}{0pt}
\setlength{\parskip}{0pt}

\bibliographystyle{IEEEbib}
\bibliography{refs}
% \clearpage
\setcounter{page}{1}
% \maketitlesupplementary

%%%%%%%%%%%%%%%%%%%%%%%%%%%
\section{supplementary}

In this supplementary material, we provide detailed implementation details, additional analyses, limitations, and qualitative results as follows:

\begin{compactitem}
\item In \secref{supp_data}, we describe the baseline configurations used in our experiments.
\item In \secref{inpainting_mask}, we demonstrate how our method enables flexible rate control without compromising performance.
\item In \secref{supp_qualitative}, we provide additional qualitative results, including comparisons on visual question answering (VQA), text-to-image (T2I) generation, text-guided image inpainting, and outpainting tasks.
\end{compactitem}

\subsection{Details of Comparisons with Baselines}
\label{sec:supp_data}
For fair comparison, we configure traditional codecs (BPG~\cite{bellard2015bpg}, VVC~\cite{bross2021overview}) and generative codecs (MS-ILLM~\cite{ms-illm}, VQ-Kmeans~\cite{mao2024dcc}, DiffEIC~\cite{diffeic}) under settings that yield bitrates comparable to UniMIC across different tasks. Specifically:

\begin{enumerate}
    \item \textbf{Text-to-Image Generation:}  
    BPG uses HEVC-Intra 18.0 with QP = 51, while VVC adopts VVC-Intra 11.0 with QP = 52.
    Generative codecs are tested with publicly available pretrained models:  
    MS-ILLM with \texttt{msillm\_quality\_1}, VQ-Kmeans with a 2048-entry codebook, and DiffEIC with the \texttt{1\_2\_4} configuration.
    
    \item \textbf{Text-Guided Image Inpainting:}  
    BPG is evaluated at QP = 50 (edge-to-cloud) and QP = 51 (cloud-to-edge), while VVC is evaluated at QP = 52 and QP = 69, respectively.  
    For generative codecs, MS-ILLM uses \texttt{msillm\_quality\_1} (edge-to-cloud) and \texttt{msillm\_vlov1\_1} (cloud-to-edge); VQ-Kmeans employs a 2048-entry codebook for edge-to-cloud and a 256-entry codebook for cloud-to-edge; DiffEIC is tested with \texttt{1\_2\_8} and \texttt{1\_2\_16}, respectively.
    
    \item \textbf{Text-Guided Image Outpainting:}  
    BPG is evaluated at QP = 47 (edge-to-cloud) and QP = 46 (cloud-to-edge), while VVC is evaluated at QP = 47 and QP = 45, respectively.  
    For generative codecs, MS-ILLM employs \texttt{msillm\_quality\_1} for both directions; VQ-Kmeans uses a 2048-entry codebook (cloud-to-edge) and a 256-entry codebook (edge-to-cloud); DiffEIC is evaluated with the \texttt{1\_2\_8} configuration for both directions.
    
    \item \textbf{Visual Question Answering:}  
    BPG is evaluated at QP = 51, while VVC version 11.0 (intra coding) is evaluated at QP = 49.  
    Generative codecs are tested using MS-ILLM (\texttt{msillm\_quality\_1}), VQ-Kmeans (2048-entry codebook), and DiffEIC (\texttt{1\_2\_4}).
\end{enumerate}

\subsection{Flexible Bitrate Control}
\label{sec:inpainting_mask}
\begin{figure*}
    \flushright
    \includegraphics[width=1\linewidth]{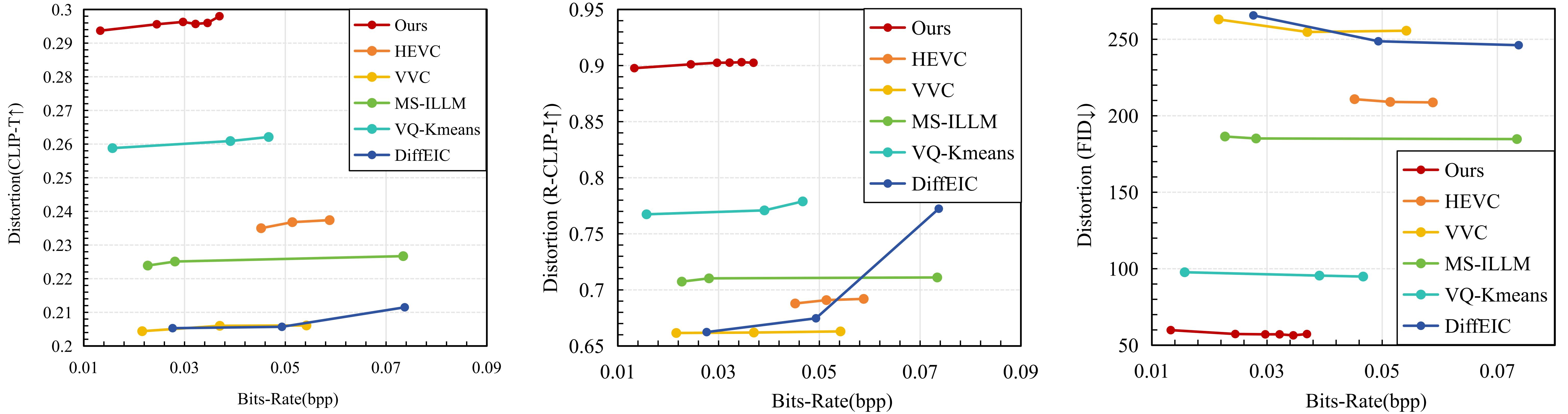}
   \caption{
   The R-D performance of BPG, VVC, MS-ILLM, VQ-Kmeans, DiffEIC, and our method on the MagicBrush dataset. (a) A higher CLIP-T score indicates better semantic consistency between the generated or edited images and their corresponding text descriptions;  (b) A higher R-CLIP-I score reflects better structural preservation between the original and edited images; (c) A lower FID score suggests improved overall image quality and a closer match between the distributions of the generated (or edited) images and real images.
   }
    %\vspace{-2mm}
    \label{fig:inpainting_mask}
\end{figure*}
The proposed Uni-MIC framework enables flexible bitrate control without the need for additional training. 
Specifically, we can mask image tokens corresponding to less informative regions of the original image based on a predefined ratio, thereby effectively reducing the transmission bitrate. 
As illustrated in the \figref{inpainting_mask}, even when most image tokens are masked during the user-to-cloud stage in the text-guided inpainting task, our method still significantly outperforms various baseline approaches in terms of CLIP-T, CLIP-I, and FID metrics.
\subsection{Additional Results}
 \begin{figure*}[!t]
    \centering
    \includegraphics[width=1.0\linewidth]{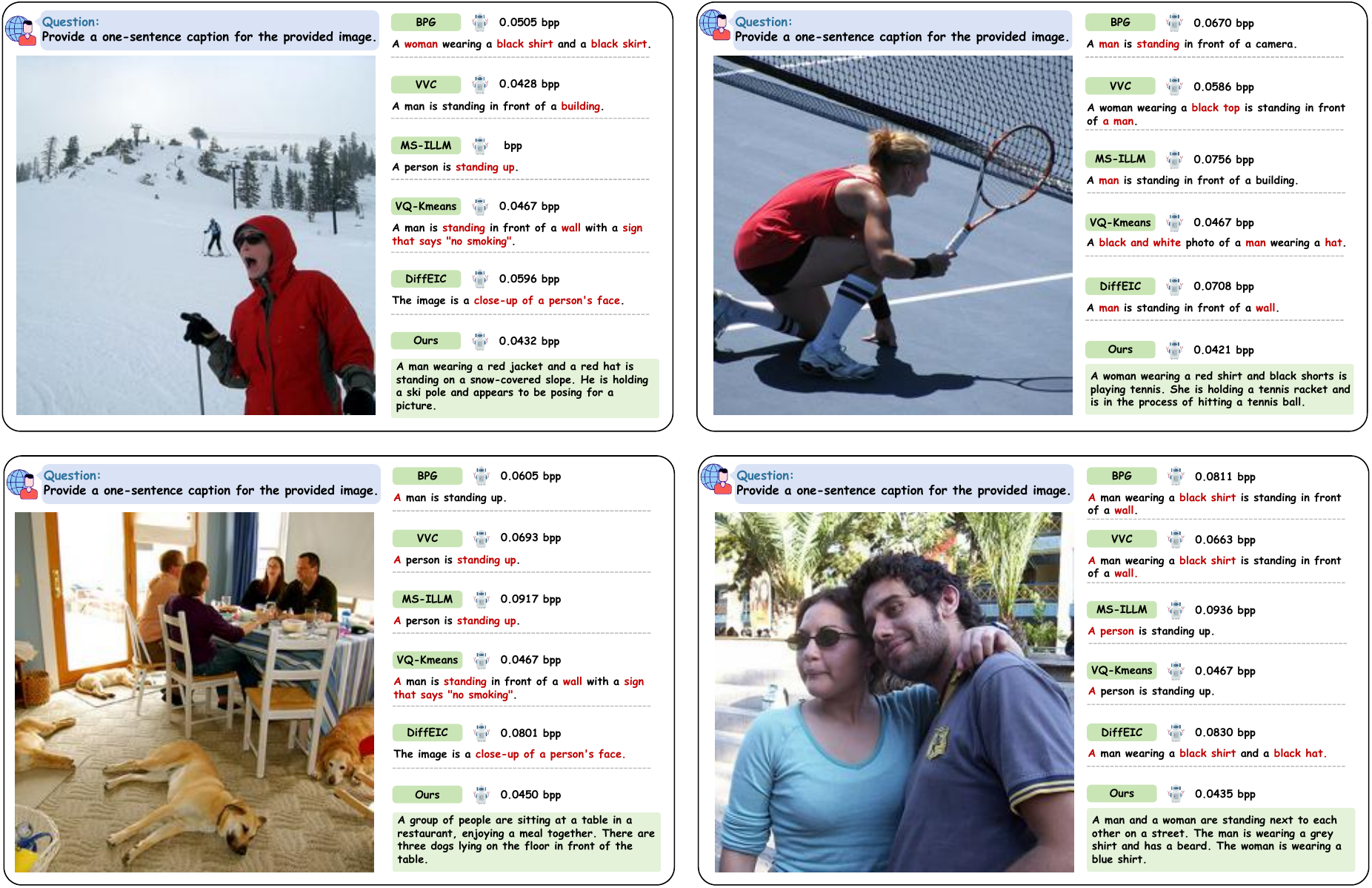}
    \caption{\textbf{Qualitative comparison of VQA capability on examples from the Flickr30k dataset~\cite{young2014image}.}  We show two representative cases with input questions and the corresponding answers generated by different methods. Baseline codecs (BPG, VVC, MS-ILLM, VQ-Kmeans, DiffEIC) either produce generic or semantically incorrect responses, whereas UniMIC generates contextually rich and accurate captions that align closely with the visual content.
    }
    \label{fig:vqa_results}
    \vspace{-3mm}
\end{figure*}

\vspace{-1mm}
\begin{figure*}[!t]
    \centering
    \includegraphics[width=0.95\linewidth]{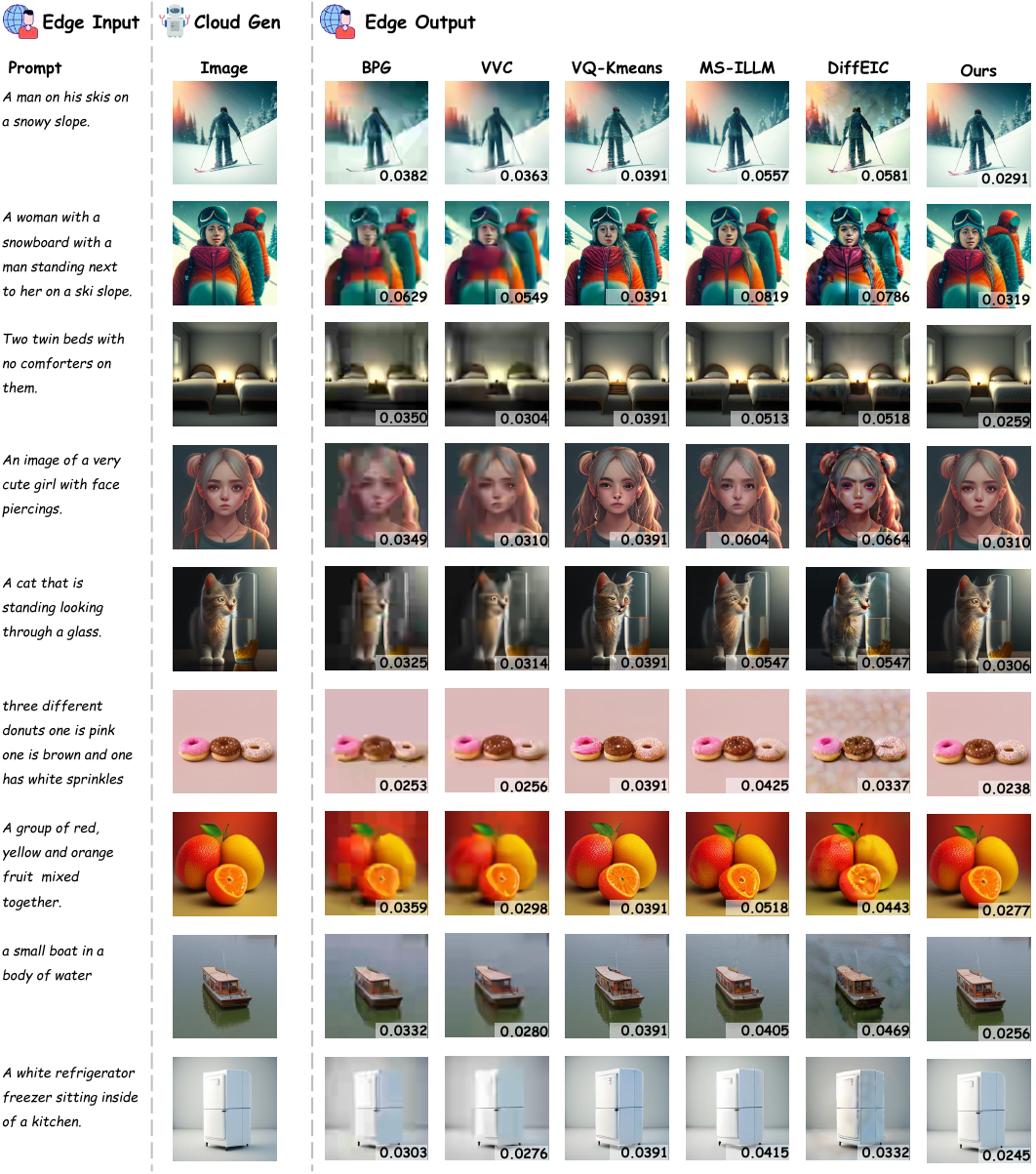}
    \caption{\textbf{Qualitative comparison of T2I generation.} 
Compared with traditional codecs and recent generative baselines, UniMIC preserves semantic details and visual fidelity while achieving lower bitrates (bpp values shown under each result).}
    \label{fig:t2i_results}
    \vspace{-5 mm}
\end{figure*}
\begin{figure*}[!t]
    \centering
    \includegraphics[width=0.95\linewidth]{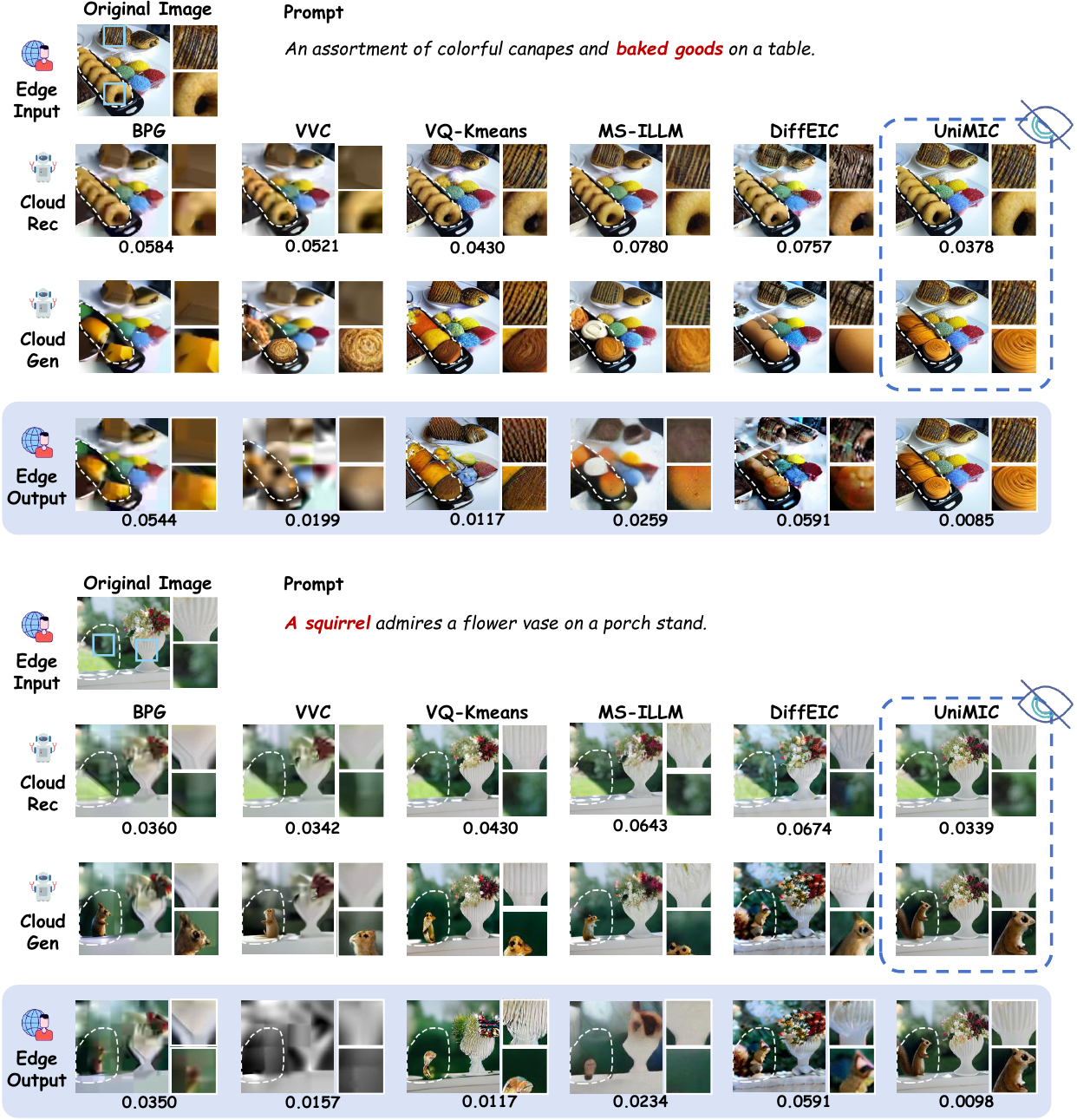}
    \caption{\textbf{Quantitative comparison results on text-guided image inpainting.}}
    \label{fig:inpainting_results}
    \vspace{-5 mm}
\end{figure*}
\label{sec:supp_qualitative}
\begin{figure*}[!t]
    \centering
    \includegraphics[width=0.95\linewidth]{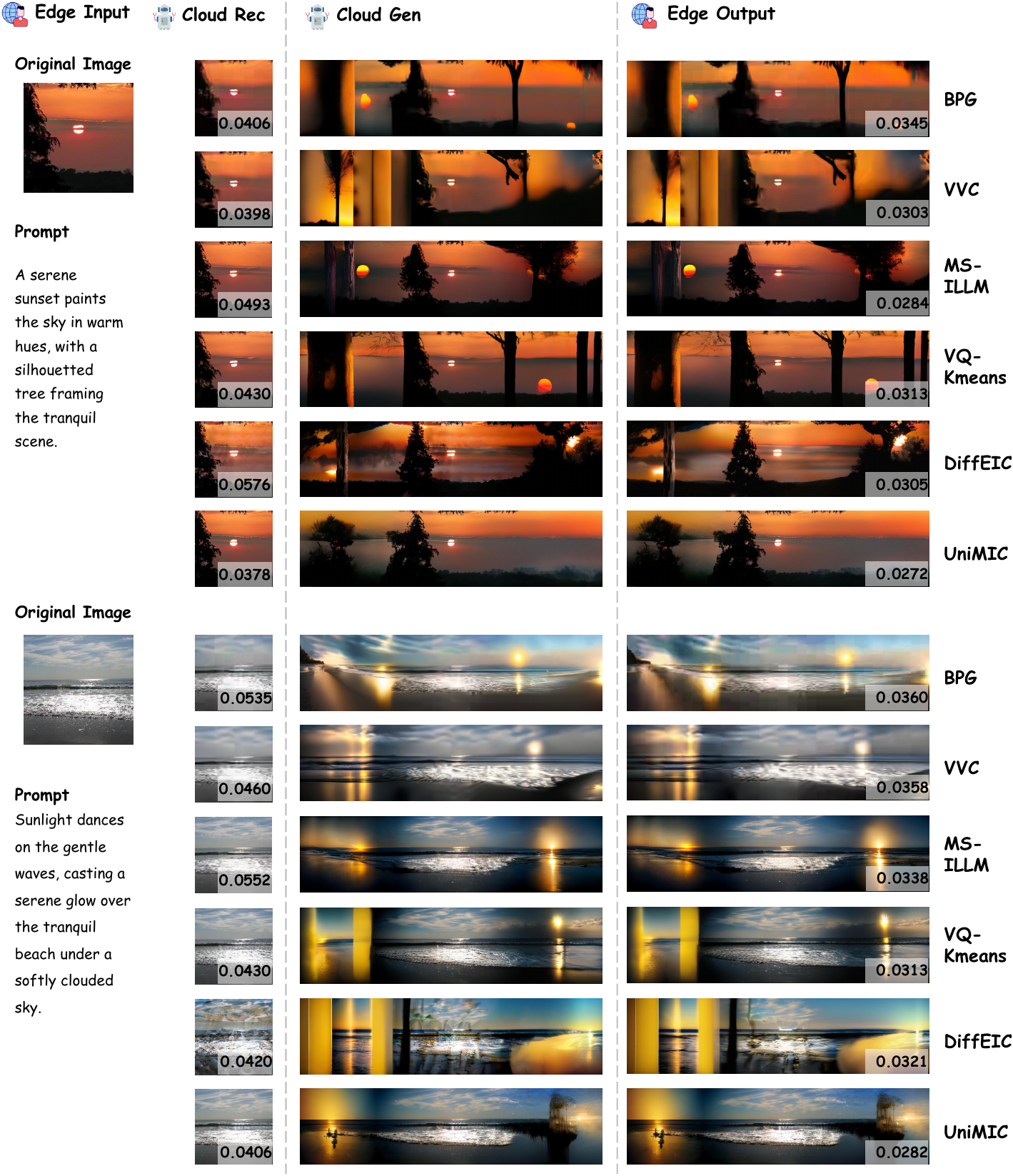}
    \caption{\textbf{Qualitative comparison results of text-guided image outpainting.} Specifically, the encoding method used to generate each image is indicated above, and the white box at the bottom denotes the decoding bitrate (bits per pixel, bpp) for each image. The red box highlights the reconstructions of the original image at different stages.}
    \label{fig:outpainting_results}
    \vspace{-5 mm}
\end{figure*}

\figref{vqa_results}, \figref{t2i_results}, \figref{inpainting_results}, and \figref{outpainting_results} provide further visual comparisons between UniMIC and baseline methods on VQA, T2I generation, text-guided image inpainting, and outpainting tasks.

\end{document}